\documentclass[a4paper, 11pt]{article}
\usepackage{times}
\usepackage{natbib}
\usepackage{mathrsfs,amssymb}
\usepackage{amsmath,color}

\usepackage{url}
\usepackage[T1]{fontenc}
\usepackage{beramono}

\usepackage{hyperref}
\usepackage{xurl}

\usepackage{mathtools}
\usepackage[all]{xy}
\newcommand{\bcx}{{\bf X}}
\newcommand{\bcy}{{\bf Y}}
\newcommand{\bcz}{{\bf Z}}
\newcommand{\bcw}{{\bf W}}
\newcommand{\bb}{{\bf b}}

\newcommand{\bx}{{\bf x}}
\newcommand{\by}{{\bf y}}
\newcommand{\bz}{{\bf z}}

\newcommand{\bmu}{{\boldsymbol\mu}}
\newcommand{\bfphi}{{\boldsymbol\varphi}}
\newcommand{\bfeta}{{\boldsymbol\eta}}
\newcommand{\beps}{{\boldsymbol\epsilon}}
\newcommand{\btheta}{{\boldsymbol\theta}}
\newcommand{\bfzeta}{{\boldsymbol\zeta}}
\newcommand{\bfsigma}{{\boldsymbol\sigma}}

\newcommand{\R}{\mathbb{R}}
\title{Deep dynamic modeling with just two time points: Can we still allow for individual trajectories?}

\author{%
	Maren Hackenberg$^{1, }$\footnote{Corresponding author: e-mail: maren.hackenberg@imbi.uni-freiburg.de}%
	\and Philipp Harms$^2$%
	\and Michelle Pfaffenlehner$^1$%
	\and Astrid Pechmann$^3$%
	\and Janbernd Kirschner$^{3,4}$ %
	\and Thorsten Schmidt$^2$%
	\and Harald Binder$^1$
}

\date{%
	$^1$ Institute of Medical Biometry and Statistics, Faculty of Medicine and Medical Center, University of Freiburg, Germany \\%
	$^2$ Institute of Mathematics, Faculty of Mathematics and Physics, University of Freiburg, Germany\\%
	$^3$ Department of Neuropediatrics and Muscle Disorders, Faculty of Medicine and Medical Center, University of Freiburg, Germany\\%
	$^4$ Department of Neuropediatrics, University Hospital Bonn, Bonn, Germany\\[2ex]%
	\today
}

\begin{document}

\maketitle

\def\spacingset#1{\renewcommand{\baselinestretch}%
	{#1}\small\normalsize} \spacingset{1}

\begin{abstract}
	Longitudinal biomedical data are often characterized by a sparse time grid and
	individual-specific development patterns.
	Specifically, in epidemiological cohort studies and clinical registries we are facing the question of what can be learned from the data in an early phase of the study, when only a baseline characterization and one follow-up measurement are available. Inspired by recent advances that allow to combine deep learning with dynamic modeling, we investigate whether such approaches can be useful for uncovering complex structure, in particular for an extreme small data setting with only two observations time points for each individual.
	Irregular spacing in time could then be used to gain more information on individual dynamics by leveraging similarity of individuals.
	We provide a brief overview of how variational autoencoders (VAEs), as a deep learning approach, can be linked to ordinary differential equations (ODEs) for dynamic modeling, and then specifically investigate the feasibility of such an approach that infers individual-specific latent trajectories by including regularity assumptions and individuals' similarity. 
	We also provide a description of this deep learning approach as a filtering task to give a statistical perspective.
	Using simulated data, we show to what extent the approach can recover individual trajectories from ODE systems with two and four unknown parameters and infer groups of individuals with similar trajectories, and where it breaks down. The results show that such dynamic deep learning approaches can be useful even in extreme small data settings, but need to be carefully adapted.
\end{abstract}

\noindent%

\spacingset{1.0}

\section{Introduction}
\label{sec:intro}

When modeling longitudinal biomedical measurements from individuals, such as in an epidemiological cohort or a clinical registry, we seek to capture the individual-specific underlying dynamic processes, e.g., corresponding to a latent health status, to extract person-specific disease trajectories. Often, however, such data are observed irregularly in time and at a small number of time points. When setting up new cohorts or registries, a typical situation even is that only baseline measurements and a single further measurement time point are available for each individual. Such an extreme setting with only two time points might seem hopeless for dynamic modeling. Yet, irregular spacing of measurements might nevertheless enable modeling of trajectories: If the second measurements of similar individuals occur at different time points, then these aggregated measurements are informative about each individual trajectory. Such similarity might be determined based on some additional baseline characteristics. The challenge then is to identify a latent representation of the longitudinal measurements where similarity can be described by some function of the additional baseline characteristics, and to incorporate a dynamic model for describing individual trajectories and quantifying similarity. We investigate how a deep learning approach, specifically based on variational autoencoders \citep[VAEs,][]{Kingma2014} for latent representations, could be useful in such a setting, and how far it can be pushed in terms of model complexity, before it breaks down due to the limited amount of information. 
Deep learning approaches have been shown to be successful at uncovering complex structure in data. They have recently been combined with dynamic modeling techniques, such as ordinary differential equations (ODEs), promising to combine the advantages of explicitly incorporating structural assumptions with data-driven approaches \citep{Chen2018, Brouwer2019, Yildiz2019, Rackauckas2020}.

We investigate whether individual-specific latent development patterns can be identified by such approaches in our setting. To this end, we develop an approach that incorporates both dynamic modeling by differential equations and regularity assumptions regarding individuals' similarity. This dynamic modeling is performed at the level of a latent representation obtained from a VAE, a generative deep learning approach based on the statistical framework of variational inference \citep{Wand2010}. We impose an ODE system on the latent space and 
use implicit regularity assumptions on individuals' similarity to fit individual-specific ODE parameters conditional on additional baseline variables. 
We further leverage individuals' similarity in an extension of the approach, where we enrich each individual's information by assigning it to a group of similar ones and weight individuals' contribution to the overall loss function according to their similarity. 
These concepts can be viewed as a transfer of the principles underlying localized regression, which allow for person-specific linear regression models and assign weights to the contribution of individuals to the likelihood used for estimation \citep{Loader1999, Tutz2005}. To provide an additional statistical perspective, we discuss how the approach can be formulated as a filtering problem \citep{Harvey1989, West1989} with individually parametrized dynamics. 

Several methods exist to model latent dynamics by differential equations within a neural network framework, but most rely on a fixed time grid with dense, regularly spaced observations \citep{Chen2018, Yildiz2019, Rackauckas2020}. While \citet{Rubanova2019} propose an extension for irregularly sampled data, here also many time points are required. \citet{Brouwer2019} address modeling of a sporadically observed time series and also discuss a filtering perspective, but their approach does not infer a compact latent representation and is applicable only to low-dimensional systems observed with little noise.
Other techniques model dynamic processes within a VAE architecture without ODEs \citep[e.g.,][]{Gregor2018, Fortuin2019}, but do not explicitly capture the underlying trajectories, thus not allowing for inter- or extrapolation of the identified trajectory. In contrast, the approach that we investigate is explicitly built to obtain trajectories to gain knowledge on disease processes.

From a more general perspective, algorithmic models in machine learning, such as neural networks, have been viewed as distinct from explicit mathematical or classical statistical models \citep{Breiman2001} as they treat the data-generating mechanism as a black box and rely predominantly on prediction performance. Being increasingly complex and heavily parametrized, such models typically require large amounts of training data for optimization and lack the interpretability of an explicit data model \citep{Breiman2001, Innes2019_Zygote_dP}. Recent research implies that such black-box models can therefore greatly benefit from including explicit modeling components like differential equations, representing a shift towards simpler models that take advantage of problem structure and can be trained with less data \citep{Chen2018, Rackauckas2020}. The approach that we investigate builds on this idea, and we explore how far this concept of combining deep learning with modeling carries in an extreme small data setting with only two observation time points. 

In the following, we give an overview of the general data structure and modeling challenge, to characterize the application settings where our results might be useful. Next, we provide a brief introduction to variational inference for latent representations in combination with ODEs from a biostatistician's point of view.
We then present the detailed approach for extracting individual-specific development patterns, discuss it from a filtering perspective and describe the estimation procedure, including how it can be extended to leverage individuals' similarity. We show results from a simulation study in various scenarios with different underlying ODE systems, also to explore where the approach breaks down due the limited amount of information. Additionally, we illustrate the feasibility of the approach in a real application with a recently set up rare disease registry. Finally, we discuss our findings, pointing out limitations of the approach and directions for future research.

\section{Application setting and modeling challenge}
\label{sec:applicationsetting}

Our investigation into the usefulness and limits of combining deep learning with dynamic modeling has originally been motivated by a scenario from a sub-study of the NAKO, a large German epidemiological cohort study.
For each individual, a large number of measurements was taken at a baseline time point, but only a small subset of these measurements was repeated at a subsequent time point.
This subsequent time point varies between individuals, resulting in a very sparse (only two time points) and irregular time grid.
This structure is typical for many data sets beyond the NAKO example, including for example registries of rare diseases. As an illustrative example, we present an application on such a data set, specifically a registry of patients with spinal muscular atrophy, in Section \ref{sec:SMArtCAREapplication}. 

\begin{figure}[htb]
	\begin{center}
		\includegraphics[width=1\linewidth]{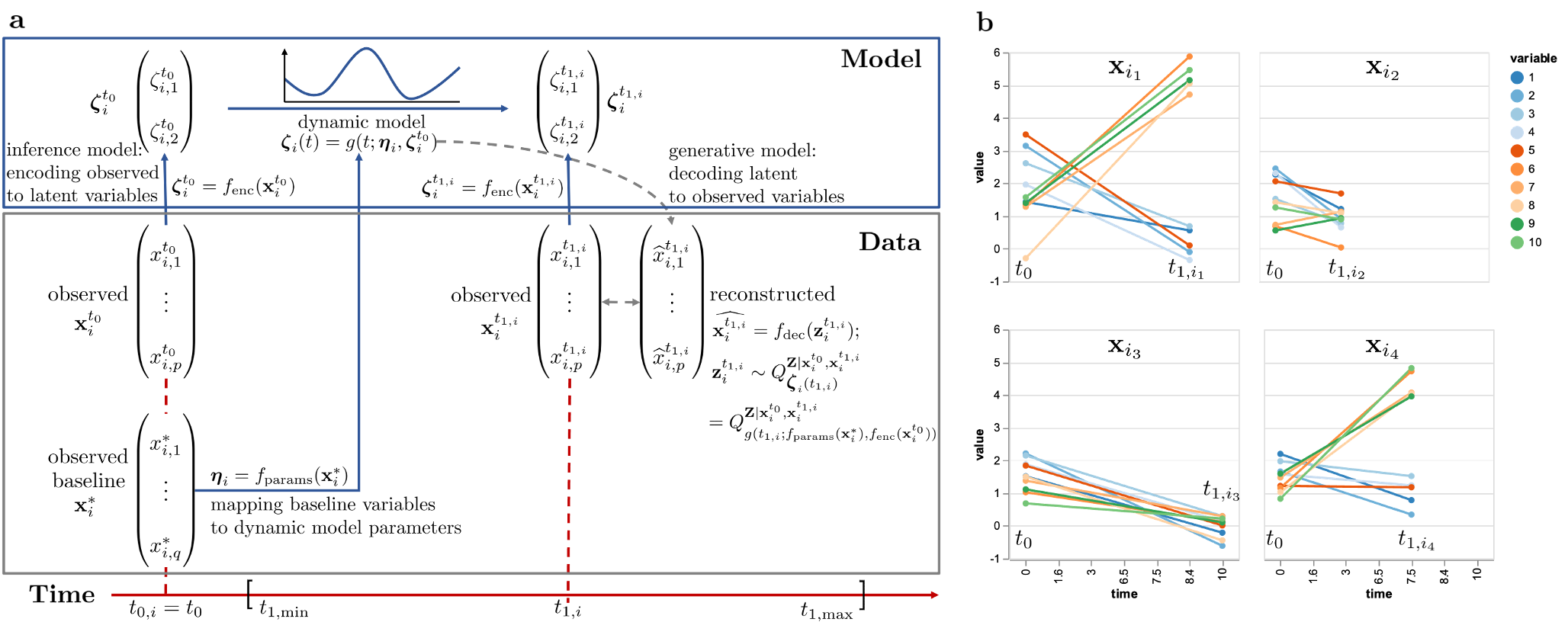}
		\caption{\textbf{a}: Schematic overview of data setting and model. \textbf{b}: Exemplary simulated data. Each panel represents one individual, dots represent individual measurements and colors indicate different measured variables. The $x$-axis denotes the time of measurements and the $y$-axis their values.}
		\label{fig:schematic_overview}
	\end{center}
\end{figure}

A schematic overview of the data structure we have in mind and the corresponding modeling challenge is presented in Figure~\ref{fig:schematic_overview}.\textbf{a}: For each individual $i$, we have two measurements $\bx_i^{t_{0,i}}$ and $\bx_i^{t_{1,i}}$, each one consisting of $p$ variables. The measurements are taken at individual-specific times $t_{0,i}$ and $t_{1,i}$, respectively. In the case of the NAKO data, the initial measurement times $t_{0,i}$ are the same for all individuals, i.e., $t_{0,i}= t_0$ for all $i$. Additionally, measurements $\bx^*_i$ of $q$ baseline variables are available at the baseline time point $t_0$ only. 
We thus assume to have baseline measurements at a common reference time point $t_0$, that corresponds to, e.g., the time point of study enrollment. Accordingly, we consider the time point of the baseline measurement as fixed reference time point $t_0$ for all individuals that is pre-specified by the specific application scenario.

Figure~\ref{fig:schematic_overview}.\textbf{b} exemplarily illustrates such structure with data from a simulation design that later on will also be used for evaluation. The time point of the second measurement varies as contrasted by individuals ${i_1}$ and ${i_2}$. Distinct development patterns are reflected by two groups of individuals with different trends (e.g., individual ${i_1}$ vs. individual ${i_3}$), and the observations are characterized by groups of variables that share common trends, suggesting an underlying lower-dimensional structure. More precisely, we assume that the measurements $\bx_i^{t_{0,i}}$ and $\bx_i^{t_{1,i}}$, as well as their timings $t_{0,i}$ and $t_{1,i}$, are determined by low-dimensional latent variables $\bz_i^{t_{0,i}}$ and $\bz_i^{t_{1,i}}$.
This corresponds to the idea that observed quantities jointly reflect some underlying health status or disease process.

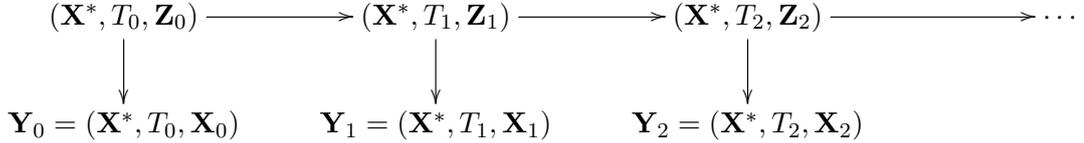
\begin{figure}[htb]
	\begin{equation*}
	\xymatrix@!C{
		(\bcx^*,T_0,\bcz_0) \ar[r] \ar[d] 
		&
		(\bcx^*,T_1,\bcz_1) \ar[r] \ar[d]
		&
		(\bcx^*,T_2,\bcz_2)\ar[r] \ar[d]
		&
		\cdots
		\\
		\bcy_0=(\bcx^*,T_0,\bcx_0)
		&
		\bcy_1=(\bcx^*,T_1,\bcx_1) 
		&
		\bcy_2=(\bcx^*,T_2,\bcx_2)}
	\end{equation*}
	\caption{Hidden Markov structure of the data and model. The observed variables are the baseline measurement $\bcx^*$, the measurement times $T_k$, and the measurements $\bcx_k$. The unobserved variables are the latent states $\bcz_k$, whose dynamics depend on the baseline measurements $\bcx^*$, i.e., $(\bcx^*,T_k,\bcz_k)_{k=1,\dots,K}$ is a Markov process.}
	\label{fig:hmm}
\end{figure}

While we will address the corresponding modeling challenge with approaches suggested by the deep learning community, it is important to note that the problem can also be thoroughly described from a statistical perspective, regardless of the estimation approach that is chosen later on.
The probabilistic dependence between the variables is well described by a hidden Markov model \citep{cappe2006inference}, whose dependence structure has the graphical representation in Figure~\ref{fig:hmm}. 
There and elsewhere in this paper, we denote random variables by capital letters and their realizations with small letters, using boldface to distinguish vectors from scalars.  
Accordingly, $\bx^{t_{0,i}}_i$ are realizations of $\bcx_0$, $\bx^{t_{1,i}}_i$ are realizations of $\bcx_1$, $\bz^{t_{0,i}}_i$ are realizations of $\bcz_0$, $\bz^{t_{1,i}}_i$ are realizations of $\bcz_1$, $t_{0,i}$ are realizations of $T_0$, and $t_{1,i}$ are realizations of $T_1$, for each $i=1,\dots,n$. 
The model easily extends to $K\geq 2$ time points, as shown in Figure~\ref{fig:hmm}. Yet, we focus on the extreme setting of just two time points.
Each arrow in the diagram stands for a probability kernel, which determines the conditional distribution of the corresponding random variables.
In particular, the diagram implies that the process $(\bcx^*,T_k,\bcz_k)_{k=1,\dots,K}$ is Markovian.
Most importantly, from a modeling perspective, this implies that the dynamics of the latent state $(\bcz_k)_{k=1,\dots,K}$ depend on the baseline variable $\bcx^*$, and also on time. This is plausible, as health (and disease processes) will to some extent depend, e.g., on life style and/or disease characteristics that can be determined at baseline.

Probabilistic inference in hidden Markov models is commonly referred to as filtering \citep{cappe2006inference}. 
More precisely, filtering refers to the computation of the conditional distribution of the unobserved variable $\bcz_k$ given the observations $\bcy_1,\dots,\bcy_k$ at an instant $k\leq K$. Smoothing refers to the computation of the conditional distribution of $\bcz_k$ given $\bcy_k$ for $k\leq K$. We are interested in a combination of both, namely, the computation of the conditional distribution of $\bcz_1,\dots,\bcz_k$ given $\bcy_1,\dots,\bcy_k$, as this is needed for prediction of the observation process $(\bcy_t)_{t\geq 0}$ in between and beyond the observation times $T_k$, $k=1,\dots, K$.
We will refer to this task simply as a filtering task.
It is complicated by the fact that the hidden Markov structure is unknown, i.e., the probability kernels depicted by arrows in Figure~\ref{fig:hmm} are unspecified.
The maximum likelihood estimation of these kernels can be achieved by variational inference, which is briefly reviewed in general terms in Section~\ref{sec:background} and described in application-specific terms in Section~\ref{subsec:filtering}.

In the specific approach that we will consider in the following, the measurements $\bcx_k$ are encoded into the low-dimensional representation $\bcz_k$. 
Algorithmically, this is achieved by mapping the observations $\bcx_k$ to certain distribution parameters $\bfzeta_k$ of the random variable $\bcz_k$ (Figure~\ref{fig:schematic_overview}.\textbf{a}). 
Thus, in terms of the realizations of these random variables, we set $\bfzeta_i^{t_0} = f_{\mathrm{enc}}(\bx_i^{t_0})$ and $\bfzeta_i^{t_{1,i}} = f_{\mathrm{enc}}(\bx_i^{t_{1,i}})$. 
To obtain an individually parametrized model of the dynamics of the latent representation, we map the additional baseline variables $\bx^*_i$ to a distinct set of dynamic model parameters $\bfeta_i=f_{\mathrm{params}}(\bx^*_i)$. 
We can thus describe the individual-specific latent dynamics as a function of $t$ by defining $\bfzeta_i(t) = g(t, \bfeta_i, \bmu_i^{t_0}) = g(t, f_{\mathrm{params}}(\bx^*_i), f_{\mathrm{enc}}(\bx_i^{t_0}))$, where the function $g$ describes the dynamic model conditional on the individual-specific parameters obtained from baseline variables and the latent representation at the baseline time point $\bfzeta_i^{t_0}$. We then define a mapping from the latent space back to the original data space to obtain a reconstructed observation $\widehat{\bx_i}^{t_{1,i}}$. Specifically, we build a generative model and draw $\bz^{t_{1,i}}$ from the conditional distribution $Q_{\bfzeta(t_1)}^{Z\mid \bx_i^{t_0}, \bx_i^{t_{1,i}}}$ of the latent variable with mean $\bfzeta$ following the smooth dynamic defined by $g$ and given the original data $\bx_i^{t_0}, \bx_i^{t_{1,i}}$, to obtain a reconstructed observation $\widehat{\bx_i}^{t_{1,i}} =f_{\mathrm{dec}}(\bz^{t_{1,i}})$.

\section{Theoretical background}
\label{sec:background}

In the following, we briefly describe the framework of \emph{variational inference}, which provides the basis for VAEs, before showing how deep neural networks, which also are a component of VAEs, can be combined with differential equations for dynamic modeling. For a more comprehensive overview of variational inference from a statistical perspective see \citet{Blei2017}, or \citet{Wand2010} for a detailed tutorial.

\subsection{Variational autoencoders}

For ease of notation, we use generic vectors $\bcx$ and $\bcz$ in this section, whereas the later application involves vectors of the form $\bcy_1, \dots, \bcy_k$ and $\bcz_1,\dots,\bcz_k$ for $k=1,\dots,K$.
We denote densities of random variables by small letters with the random variable as subscript, e.g., $p_\bcz(\cdot)$ is the density of $\bcz$. We abbreviate conditional distributions and densities by writing, e.g., $p_{\bcx\mid \bz}(\bx)$ for $p_{\bcx \mid \bcz =\bz}(\bx)$, and write $\mathrm{E}_p[q]$ for the integral $\int_{-\infty}^{\infty}q(\bx)\cdot p(\bx)~d\bx$.

We assume that the data are given as realizations of a random vector $\bcx$, which originates from a typically lower-dimensional random variable $\bcz$ with prior distribution $P^\bcz$ as described by the conditional distribution $P^{\bcx \mid \bz}$. 
Having only access to the realizations $\bx$ but not the corresponding latent samples $\bz$ implies that we need to compute the posterior $P^{\bcz \mid \bx}$. Here, applying Bayes' rule 
\begin{equation*}
p_{\bcz \mid \bx}(\bz) = \frac{p_{\bcx,\bcz}(\bx, \bz)}{p_{\bcx}(\bx)} = \frac{p_{\bcx,\bcz}(\bx, \bz)}{\int p_{\bcx,\bcz}(\bx, \bz) d\bz}
\end{equation*}
shows that computing the posterior requires integrating over all possible values of $\bz$. This tends to be computationally intractable in sufficiently complex Bayesian models including latent variables. In this context, variational inference provides a viable framework for approximating the posterior. The key idea is to reformulate the problem as an optimization rather than a sampling problem \citep{Blei2017}. To this end, a family $\mathcal{Q}$ of densities over the latent variables $\bcz$ is defined to find the member of the family that minimizes the \emph{Kullback-Leibler (KL) divergence} $D_{\mathrm{KL}}(q \Vert p) := \mathrm{E}_{q}\left[\log\left(\frac{q}{p}\right)\right]$ to the exact posterior,
\begin{equation*}
q^*_{\bcz \mid \bx} = \arg \min_{q_{\bcz \mid \bx}\in\mathcal{Q}} D_{\mathrm{KL}}(q_{\bcz \mid \bx}\Vert p_{\bcz \mid \bx}) = \arg \min_{q_{\bcz \mid \bx}\in\mathcal{Q}} \mathrm{E}_{q_{\bcz \mid \bx}}[\log(q_{\bcz \mid \bx}) - \log(p_{\bcx,\bcz}(\bx,\cdot))] + \log(p_{\bcx}(\bx)).
\end{equation*} 
Since $\log(p_{\bcx}(\bx))$ is a constant with respect to $q_{\bcz \mid \bx}$, minimizing $\mathrm{E}_{q_{\bcz \mid \bx}}[\log(q_{\bcz \mid \bx})] - \mathrm{E}_{q_{\bcz \mid \bx}}[\log(p_{\bcx,\bcz}(\bx,\cdot))]$ is equivalent to minimizing $D_{\mathrm{KL}}(q_{\bcz \mid \bx}\Vert p_{\bcz \mid \bx})$ with respect to $q_{\bcz \mid \bx}$. 
Since the KL-divergence is non-negative, we can derive a lower bound on the marginal data likelihood $\log(p_{\bcx}(\bx))$ called the \emph{evidence lower bound (ELBO)}:
\begin{equation}\label{eq:elbo-definition}
\begin{split}
\log(p_{\bcx}(\bx)) &= \mathrm{E}_{q_{\bcz \mid \bx}}[\log(p_{\bcx,\bcz}(\bx, \cdot))] - \mathrm{E}_{q_{\bcz \mid \bx}}[\log(q_{\bcz \mid \bx})] + D_{\mathrm{KL}}(q_{\bcz \mid \bx}\Vert p_{\bcz \mid \bx})\\
&\geq \mathrm{E}_{q_{\bcz \mid \bx}}[\log(p_{\bcx,\bcz}(\bx, \cdot))] - \mathrm{E}_{q_{\bcz\mid \bx}}[\log(q_{\bcz \mid \bx})] =: \mathrm{ELBO}(\bx, q_{\bcz \mid \bx}).
\end{split}
\end{equation}
Thus, minimizing the KL-divergence $D_{\mathrm{KL}}(q_{\bcz \mid \bx}\Vert p_{\bcz \mid \bx})$ is equivalent to maximizing the ELBO and 
$$
q_{\bcz \mid \bx}^* = \arg\min\limits_{q_{\bcz \mid \bx}\in\mathcal{Q}}\, D_{\mathrm{KL}}(q_{\bcz \mid \bx}\Vert p_{\bcz \mid \bx}) = \arg\min\limits_{q_{\bcz \mid \bx}\in\mathcal{Q}}\, \mathrm{ELBO}(\bx, q_{\bcz \mid \bx}).
$$
To gain some intuition about the properties of the optimal variational density, we can rewrite the ELBO as $E_{q_{\bcz \mid \bx}}[\log(p_{\bcx \mid \bcz}(\bx))] - D_{\mathrm{KL}}(q_{\bcz \mid \bx}\Vert p_{\bcz})$, where the first term shows that maximizing the ELBO encourages densities placing mass on configurations of latent variables that explain the observed data, while the second term has a regularizing effect and encourages densities close to the prior. 

We parametrize the prior distribution of $\bcz$ and the distribution of $(\bcx \vert \bz)$ by $\btheta \in \Theta$ and assume a parametric variational family $\mathcal{Q} = (q_{\bcz \mid \bx}(\cdot, \bfphi))_{\bfphi \in \Phi}$ with a finite-dimensional space $\Phi \subset \R^e$, that completely characterizes the family of approximate posteriors of $(\bcz \vert \bx)$. Hence, we can view the ELBO as a function of the data points $\bx$ and the parameters $\btheta$ and $\bfphi$.
Maximizing the ELBO as a lower bound on $\log(p_{\bcx}(\bx,\btheta))$ with respect to $\btheta$ yields an approximation to the maximum likelihood estimate for $\btheta$ that is better the smaller the KL-divergence 
$D_{\mathrm{KL}}(q_{\bcz \mid \bx}(\cdot,\bfphi)\Vert p_{\bcz \mid \bx}(\cdot,\btheta))$
that determines the tightness of the bound. We can thus derive both approximate maximum likelihood estimates for $\btheta$ and an optimal variational density $q_{\bcz \mid \bx}$ by maximizing the ELBO both with respect to $\btheta$ and $\bfphi$.

While obtaining an estimate of the gradient with respect to $\btheta$ is straightforward, we have to employ a change of variables called the \emph{reparameterization trick} \citep{Kingma2014} to estimate the gradient with respect to the variational parameters $\bfphi$. Intuitively, instead of sampling $\bz \sim Q_{\bfphi}^{\bcz \mid \bx}$, we sample some $\beps$ from a random variable independent of $\bcx$ and express $\bz$ as a deterministic transformation of $\bfphi$ and $\beps$. This allows us to obtain unbiased estimates of the ELBO with respect to both $\btheta$ and $\bfphi$ and optimize it using stochastic gradient descent \citep{Kingma2019}. For this, typically a standard normal distribution is assumed as a prior over $\bcz$.

An approach that learns a low-dimensional representation of input data based on variational inference by using neural networks to parametrize the distributions $q_{\bcz \mid \bx}(\cdot,\bfphi)$ and $p_{\bcx \mid \bz}(\cdot,\btheta)$, realizing encoding and decoding of the data into a latent space, is called a \emph{variational autoencoder (VAE)} and was first proposed in \citet{Kingma2014}. Briefly, a \emph{neural network} is a function composition $f: \R^{k_1} \to \R^{k_{n+1}}, \bx \mapsto (g_n\circ g_{n-1} \circ \dots \circ g_2\circ g_1)(\bx)$ for distinct, continuous functions $g_i: \R^{k_i} \to \R^{k_{i+1}}$ called the \emph{layers} of the network. Each layer is of the form $g_i(\bx) = h_i(W_i \bx + b_i)$, where $h_i: \R \to \R$ is a continuous non-linear function called the \emph{activation function}, $\bcw_i \in \R^{k_{i+1}\times k_i}$ and $\bb_i \in \R^{k_{i+1}}$ are \emph{weights} and \emph{biases}, also called \emph{parameters}, and $h_i$ is applied element-wise. 
To obtain partial derivatives with respect to each parameter for estimation by stochastic gradient descent, the chain rule has to be applied repeatedly due to the structure of the network as function composition, resulting in the propagation of gradients "backwards" through the layers of the network \citep[hence the name \emph{backpropagation}, ][]{Rumelhart1986}. 

\subsection{Incorporating differential equations}\label{subsec:background_integratingDEs}

Recently, \citet{Chen2018} have proposed a combination of neural networks with dynamic modeling by ordinary differential equations:
When training a neural network that includes an ODE-solving step with backpropagation, one also needs to differentiate backwards through the ODE-solving step, i.e., obtain gradients of the loss function with respect to all inputs of the ODE solver. 
This is enabled by \emph{differentiable programming}, a paradigm based on the idea of using automatic differentiation frameworks to differentiate through arbitrary computer programs \citep{Innes2019_Zygote_dP}. Differentiable programming thus allows for flexible models that seamlessly integrate different building blocks from, e.g., deep neural networks and mechanistic modeling, by viewing the models as functions specified by computer programs and using automatic differentiation to provide gradients for parameter estimation. More generally, this represents a shift of neural networks towards modeling by incorporating components such as differential equations that can better reflect the problem structure in the task at hand \citep{Rackauckas2020}. 

\section{An approach for individual latent dynamics}
\label{sec:methods}

\subsection{Integrating individually parametrized latent dynamics into  a variational autoencoder framework}
\label{subsec:methodoverview}

\begin{figure}[htb]
	\includegraphics[width=\linewidth]{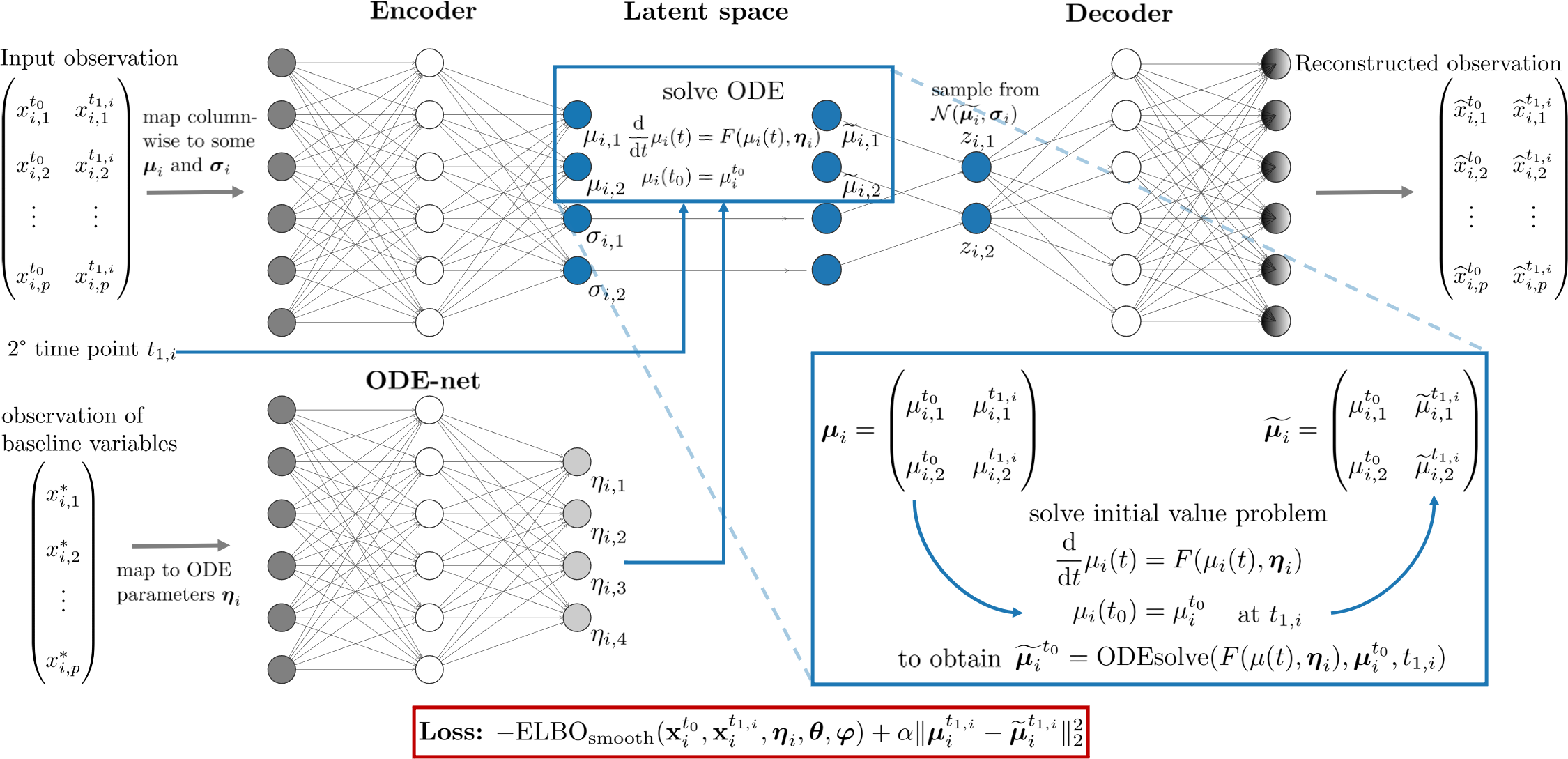}
	\caption{Overview of model architecture and training.}
	\label{fig:model_architecture}
\end{figure}	

Our suggested approach to allow for person-specific dynamics in a latent representation is schematically illustrated in Figure~\ref{fig:model_architecture}.
To extract a lower-dimensional representation underlying the data, we rely on the framework of variational inference, as outlined in Section~\ref{sec:background}, and define the functions $f_{\mathrm{enc}}$ and $f_{\mathrm{dec}}$ from Section~\ref{sec:applicationsetting} as neural networks that parametrize the distributions $q_{\bcz \mid \bx}(\cdot,\bfphi)$ and $p_{\bcx \mid \bz}(\cdot,\btheta)$, such that our model is based on a VAE architecture.

To model smooth latent trajectories, we define an ODE system that describes the latent space dynamics, i.e., the function $g$ in Section~\ref{sec:applicationsetting} corresponds to the solution of this ODE system. 
By prescribing that the posterior mean of the latent variable follows the dynamic pattern specified by the chosen ODE system, we thus impose additional structural assumptions on the latent space. This added structure has a regularizing effect as it encourages the model to find a latent representation that matches this smoothness assumption, and can thus help to prevent potential overfitting.
Instead of giving the model complete freedom in structuring its latent space, the ODE system enforces a representation that corresponds to the specific dynamical system of this ODE, thus acting as an additional constraint on the model that guides it towards a latent representation with smooth dynamics.

To model the stochasticity of the measurement process, we sample $\bz$ from the posterior distribution according to smooth dynamics after solving the ODE, i.e., from the distribution with mean given by the ODE solution at a specific time point. We then decode it to data space via the neural network $f_{\mathrm{dec}}$ to obtain a reconstructed observation $\widehat{\bx}$.

To extract individual-specific development patterns, i.e., to solve a distinct ODE system for the posterior mean of each individual observation, we employ the additional variables $\bx_i^* \in \R^q$ measured only at the baseline time point. This reflects the assumption that this more extensive characterization carries information about each individual's development. 
To allow for a flexible transformation from the baseline variables to an individual set of ODE parameters $\bfeta_i$, we define the mapping $f_{\mathrm{params}}$ as a neural network, the ODE-net. This implicitly incorporates that individuals with similar baseline characteristics exhibit similar dynamics. A differentiable programming framework allows to train this additional network jointly with the VAE and the dynamic model. As a result, the individually parametrized ODE structures can influence the model to find an appropriate latent representation matching the assumption of a smooth trajectory, and vice versa, the inference of ODE parameters is guided by the latent space dynamics. 	

\subsection{The approach from a filtering perspective}
\label{subsec:filtering}

Returning to the statistical perspective from Section 2, the variational inference described above 
can be used to solve the filtering problem.  
One specifies a parametric family of approximate distributions $q_{\bcz|\by}(\cdot,\bfphi)$ indexed by the parameter $\bfphi$. 
This family of distributions should contain (at least approximately) the posterior distribution $p_{\bcz|\by}(\cdot,\btheta)$.
If $\btheta$ is known, as in the classical filtering setting, the joint distribution $p_{\bcy,\bcz}(\cdot,\btheta)$ is given by the structure of the hidden Markov model. 
In this case all distributions which appear in the evidence lower bound are specified.
The variational filter then seeks the maximizer $q^*_{\bcz|\by}$ of the evidence lower bound, and this maximizer coincides (at least approximately) with the posterior distribution $p_{\bcz|\by}(\cdot,\btheta)$. 
Moreover, using the estimated posterior distribution $q^*_{\bcz|\by}$, the algorithm is able to interpolate and extrapolate the dynamics of the process $(\bcy_t)_{t\geq 0}$ in-between and beyond the given observation times $T_i$, which is our ultimate goal in the application. 

The far-reaching claim, which is explored in the present paper, is that meaningful prediction of the process $(\bcy_t)_{t\geq 0}$ is possible even if the parameter $\btheta$ has to be estimated. 
A well-known algorithm tackling this task is the EM-algorithm \citep{Rubin1977}, which iterates maximum likelihood estimation and filtering. 
Our variational filter solves these two problems in one step, as we explain next.
Note that the decoder parametrizes a set of candidate joint distributions $q_{\bcy,\bcz}$, each one being compatible with the dependence structure  in Figure~\ref{fig:hmm}.
Now assume that the actual joint distribution is contained in this set (or at least well approximated by some element in it). 
Then the maximization of the evidence lower bound, which is performed by our variational filter, has two effects: first, it performs a maximum likelihood search over all candidate joint distributions, and second, for the optimal joint distribution it computes the posterior distribution $p_{\bcz|\by}$.    
Thus, under these assumptions the variational filter is able to make correct predictions about $(\bcy_t)_{t\geq 0}$ despite the fact that no latent dynamics are specified. 

While filtering algorithms are designed to work well even for small data sets, the maximum likelihood estimation has difficulties with small data. 
The unknown parameter $\btheta$ describes the functional dependence of the latent state dynamics on the baseline measurements, as represented by the ODE-net of Figure~\ref{fig:model_architecture}, which shall be denoted by $\widehat{\btheta}$.
From a general viewpoint we therefore may view $\btheta$ as an element of an infinite-dimensional function space $\Theta$. 
To facilitate the identification of $\btheta$, we impose the similarity assumption that individuals with similar baseline variables  have similar latent space dynamics. 
Mathematically, this corresponds to the choice of a (highly regular) function space $\Theta$. 
Consequently, all $\btheta\in\Theta$ have accurate finite-dimensional approximations $\widehat{\btheta}$ in the sense of rate distortion theory \citep{devore1998nonlinear}. 
The algorithmic approach explored in this paper can be interpreted as the converse step, where one starts out with some finite-dimensional approximations $\widehat{\btheta}$, given by the ODE-net in Figure~\ref{fig:model_architecture}, and defines $\Theta$ as the set of functions which can be approximated at polynomial or exponential rates by $\widehat{\btheta}$ \citep{gribonval2019approximation}. 
This results in classical function spaces $\Theta$ of Besov or real analytic regularity, at least under certain assumptions on the network architecture and activation function \citep{gribonval2019approximation, guhring2020expressivity, boelcskei2019optimal}. 
While the error analysis for maximum likelihood estimation in this framework is left to future work, we explore algorithmically the implications of this similarity assumption on estimation for data with an extremely small number of time points.

\subsection{Obtaining gradients for parameter optimization}
\label{subsec:trainingdetails}

In the following section, we formalize our approach and derive an adapted version of the ELBO to define a suitable loss function. The model training process also is illustrated in Figure~\ref{fig:model_architecture}. 

The encoder neural network of the VAE maps an input observation $\bx_i = (\bx_i^{t_0} \quad \bx_i^{t_{1,i}}) \in \R^{p\times 2}$ column-wise to the mean  $\bmu_i= (\bmu_i^{t_0} \quad \bmu_i^{t_{1,i}}) \in \R^{m\times 2}$ and standard deviation $\bfsigma_i = (\bfsigma_i^{t_0} \quad \bfsigma_i^{t_{1,i}}) \in \R^{m\times 2}$ of the posterior distribution $Q_{\bfphi}^{\bcz\mid \bx_i} = \mathcal{N}(\bmu_i,\bfsigma_i^2)$, where $m$ is the dimension of the latent space.
Note that $Q_{\bfphi}^{\bcz\mid \bx_i}$ does not explicitly depend on time or the baseline measurements; this is modeled via latent dynamics.

The latent space dynamics are given by a pre-specified function $F(\bmu_i(t), t, \bfeta_i)$. This function that corresponds to the assumed underlying dynamical system should be chosen carefully. Depending on the specific application setting, domain expert knowledge on the dynamics that are expected to underlie the observed process can be employed. Generally, the chosen structure should be as simple as possible as a too complex system can be a source of potential overfitting. In our simulation study and application, we exemplarily consider the use of either a homogeneous two-dimensional linear ODE system or a non-linear Lotka-Volterra ODE system. For the chosen definition of $F$, we solve the initial value problem 
\begin{equation*}
\frac{d}{dt} \mu_i(t) = F(\mu_i(t), t, \bfeta_i); \quad \mu_i(t_0) = \bmu_i^{t_0}
\end{equation*}
at the time point $t_{1,i}$ of the individual's second measurement. \\
We obtain a vector $\widetilde{\bmu_i}^{t_{1,i}} = \mathrm{ODESolve}(F(\mu_i(t), t, \bfeta_i), \bmu_i^{t_0}, t_{1,i}) \in \R^m$ and define $\widetilde{\bmu_i} := (\bmu_i^{t_0} \quad \widetilde{\bmu_i}^{t_{1,i}})$ as the mean of the posterior distribution constrained to a smooth dynamic. Next, we draw a sample $\bz_i\sim \widetilde{Q}^{\bcz \mid \bx_i}_{\bfeta_i,\bfphi} = \mathcal{N}(\widetilde{\bmu_i}, \bfsigma_i^2)$ from the approximate posterior by using the reparameterization trick, i.e., $\bz_i  = \widetilde{\bmu_i} +  \bfsigma_i^2 \odot \beps$ where $\beps \sim \mathcal{N}(0,1)$ and $\odot$ denotes the element-wise product. 

The decoder neural network, that parametrizes the conditional distribution $P_{\btheta}^{\bcx \mid \bz_i}$ of the data given a sample of the posterior, maps $\bz_i = (\bz_i^{t_0}, \bz_i^{t_{1,i}})$ to a reconstructed observation $\widehat{\bx_i} = (\widehat{\bx_i}^{t_0}, \widehat{\bx_i}^{t_{1,i}})$ as a sample from $P_{\btheta}^{\bcx \mid \bz_i}$. 
Equivalently, the decoder outputs $P_{\btheta}^{\bcy \mid \bz_i}$ because $t_{1,i}$ and $\bx^*_i$ are fully observed.

To define a training objective for jointly optimizing the VAE model constrained to smooth latent space dynamics and the ODE-net,
we adapt the ELBO of \eqref{eq:elbo-definition}: To learn the latent space dynamics given by the ODE system jointly with the ODE-net, the posterior mean as obtained from solving the ODE is used. Thus, we introduce dependence of the ELBO on the ODE parameters and provide feedback for learning the weights and biases of the ODE-net: 
\begin{equation*}
\begin{split}
\mathrm{ELBO}_{\mathrm{smooth}}(\bx_i^{t_0}, \bx_i^{t_{1,i}}, \bfeta_i, \btheta, \bfphi) &= -D_{\mathrm{KL}}(\widetilde{q}_{\bcz \mid \bx_i^{t_0}, \bx_i^{t_{1,i}}}(\cdot, \bfeta_i, \bfphi) \Vert p_{\bcz}(\cdot,\btheta)) \\ &+ \mathrm{E}_{\widetilde{q}}[\log(p_{\bcx \mid \bz_i^{t_0}, \bz_i^{t_0}}(\bx_i^{t_{1,i}}, \bx_i^{t_0},\btheta))],
\end{split}
\end{equation*}
where in the expectation, we use $\widetilde{q}$ as an abbreviation for $\widetilde{q}_{\bcz \mid \bx_i^{t_0}, \bx_i^{t_{1,i}}}(\cdot, \bfeta_i, \bfphi)$ .
Since the ODE solution $\widetilde{\bmu_i}^{t_{1,i}}$ depends exclusively on the initial value $\bmu_i^{t_0}$ and the ODE parameters $\bfeta_i$, but not on the observations at the second time point $\bx_i^{t_{1,i}}$, we have $\widetilde{q}_{\bcz \mid \bx_i^{t_0}, \bx_i^{t_{1,i}}}(\cdot, \bfeta_i, \bfphi) = \widetilde{q}_{\bcz \mid \bx_i^{t_0}}(\cdot, \bfeta_i, \bfphi)$.
To provide feedback for the encoder weights and biases with respect to $\bx_i^{t_{1,i}}$ and improve the reconstruction of $\bx_i^{t_{1,i}}$ by the model, we add to the loss the squared Euclidean distance between $\bmu_i^{t_{1,i}}$ as obtained from directly passing $\bx_i^{t_{1,i}}$ through the encoder and $\widetilde{\bmu_i}^{t_{1,i}}$ as obtained from solving the ODE with a weighting factor of $\alpha\in[0,1]$:

\begin{equation} \label{eq:finalELBO}
\begin{split}
\mathcal{L}(\bx_i^{t_0}, \bx_i^{t_{1,i}}, \bfeta_i, \btheta, \bfphi) &= -\mathrm{ELBO}_{\mathrm{smooth}}(\bx_i^{t_0}, \bx_i^{t_{1,i}}, \bfeta_i, \btheta, \bfphi) + \alpha\Vert\bmu_i^{t_{1,i}} - \widetilde{\bmu_i}^{t_{1,i}}\Vert_2^2 \\
&= D_{\mathrm{KL}}(\widetilde{q}_{\bcz \mid \bx_i^{t_0}, \bx_i^{t_{1,i}}}(\cdot, \bfeta_i, \bfphi) \Vert p_{\bcz}(\cdot,\btheta)) - \mathrm{E}_{\widetilde{q}}[\log(p_{\bcx \mid \bz_i^{t_0}, \bz_i^{t_{1,i}}}(\bx_i^{t_0}, \bx_i^{t_{1,i}},\btheta))]\\
&+ \alpha\Vert\bmu_i^{t_{1,i}} - \widetilde{\bmu_i}^{t_{1,i}}\Vert_2^2.
\end{split}
\end{equation}

With this, we penalize large deviations of the posterior distribution from the encoder $q_{\bcz \mid \bx_i^{t_0}, \bx_i^{t_{1,i}}}(\cdot,\bfphi)$ from the posterior distribution $\widetilde{q}_{\bcz \mid \bx_i^{t_0}}(\cdot, \bfeta_i, \bfphi)$ constrained to smooth latent space dynamics. This encourages the model to implicitly condition the ODE solution on the data from the second time point, and to encode the data from both measurement time points into a lower-dimensional representation that reflects a smooth trajectory even before solving the ODE. As mentioned before, the ODE system thus imposes a smoothness constraint on the latent representation. It has a regularizing effect by encouraging the model to structure its latent representation in a way that matches the dynamics given by the ODE system.

Based on gradients of the loss function in \eqref{eq:finalELBO} with respect to the neural network parameters, it can be optimized with a stochastic gradient descent approach.

Details on the implementation of the approach in the Julia programming language and the computational cost can be found in the appendix. Additionally, we provide the code for reproducing our analyses and Jupyter notebooks that illustrate the approach at \url{https://github.com/maren-ha/DeepDynamicModelingWithJust2TimePoints}. 

\subsection{Incorporating similarity into iterative optimization}
\label{subsec:trainingonbatches}

We now show how we can exploit individuals' similarity and simultaneously take advantage of the irregularity of the time grid to address the small number of time points available for each individual. We consider this as an additional measure to improve the training performance in particularly challenging settings. 
Specifically, this is implemented by assigning at each step of the stochastic gradient descent to each individual a batch of individuals with similar underlying development patterns.
Here, similarity is defined according to distance of the latent ODE solutions to the trajectory of the reference individual in $L^2$-norm. 
By combining all second time point measurements, information on the common dynamics can be obtained at multiple time points.
We embed this into an iterative optimization framework that alternates between the assignment of batches of similar individuals and the joint training of the ODE-VAE model, such that both the latent representation of trajectories and the group assignment are iteratively improved. The two iterative steps in this optimization scheme, which resemble an expectation maximization algorithm, are as follows: 

First, the similarity of individuals is inferred based on their current ODE solutions in latent space. For this, every individual observation $\bx_i$ is passed through the VAE encoder to obtain an approximate posterior mean $\bmu_i= (\bmu_i^{t_0} \quad \bmu_i^{t_{1,i}}) \in \R^{k\times 2}$ and variance $\bfsigma_i = (\bfsigma_i^{t_0} \quad \bfsigma_i^{t_{1,i}}) \in \R^{k\times 2}$, while the respective baseline measurements $\bx^*_i$ are passed through the ODE-net to obtain a set of individual ODE parameters $\bfeta_i$  as in Section~\ref{subsec:trainingdetails}. Next, for each individual, the resulting ODE system is solved to obtain the posterior mean according to smooth dynamics as $\mu_i(t) = \mathrm{ODEsolve}(f(\mu_i(t), t, \bfeta_i))$.
For $\mu_{i} \in L^2([t_0,T])$ and a $T \in \R$, we can calculate a symmetric distance matrix 
$D$ where the $i,j$-th entry $(d_{i,j})_{i,j=1,\dots,n} = \left(\Vert \mu_i - \mu_j\Vert_{L^2([t_0,T])}\right)_{i,j=1,\dots,n}$ denotes the distance in $L^2$-norm between the ODE solutions $\mu_i(t)$ and $\mu_j(t)$ of individuals $i$ and $j$. For a pre-defined batch size of $b$, to each individual $i$ the $(b-1)$ individuals closest to that reference individual $i$ are assigned. 
Each individual in a batch $B_i$ is weighted according to its calculated distance to the reference individual $i$ by determining weights $w_1,\dots, w_{b}$ with a tricube kernel $K$, such that 
$w_j = \frac{K(d_{i,j})}{\sum_{\lbrace k: \bx_k \in B_i \rbrace} K(d_{i,k})}$ for all $j$ with  $\bx_j \in B_i$.

Second, the model is trained on the resulting batches in order to improve the estimates of the variational parameters $\bfphi$, the model parameters $\btheta$ and the individual ODE parameters $\bfeta_i$ with respect to the training objective $\mathcal{L}(\bx_i^{t_0}, \bx_i^{t_{1,i}}, \bfeta_i,\btheta,\bfphi)$ of \eqref{eq:finalELBO}. 
For each individual in the batch, the respective loss value is calculated as described in Section~\ref{subsec:trainingdetails}, using the ODE solution at its individual-specific second measurement time point, which now serve as proxy information for more time points (one from each individual in the batch) of the reference individual. To enrich this reference individual's information and stabilize the gradient of its loss function, we derive a common loss value for the entire batch as the weighted average of individual loss values, using the weights derived from the distance matrix of the first step: 
\begin{equation*}
\mathcal{L}_{\mathrm{batch}}(B_i, (\bfeta_k)_{\lbrace k:\bx_k \in B_i \rbrace}, \btheta,\bfphi) = \sum_{ k: \bx_k \in B_i} w_k \cdot \mathcal{L}(\bx_k^{t_0}, \bx_k^{t_{1,i}}, \bfeta_k,\btheta,\bfphi), \quad i=1,\dots, n.
\end{equation*} 

\section{Simulation study}
\label{sec:simulation}

To investigate the potential and limits of the approach described above in extreme settings with a small number of observations, we perform a simulation study by defining ground-truth ODE systems and sampling from the true trajectories.
We specify the general structure of the ODE as a linear system (Sections~\ref{subsec:sim-linear2parameters} and \ref{subsec:sim-linear4parameters}) or a non-linear Lotka-Volterra system (Section~\ref{subsec:sim-nonlinear}), where both systems are defined by four parameters. In Sections~\ref{subsec:sim-linear2parameters} and \ref{subsec:sim-nonlinear}, two of these parameters are assumed as known. 
In Section~\ref{subsec:sim-linear4parameters}, all four parameters are assumed to be unknown and are estimated with the ODE-net using the approach of Section~\ref{subsec:trainingonbatches}. 

\subsection{General simulation design}
\label{subsec:sim-design}

In all settings, two distinctly parametrized underlying types of temporal development patterns are simulated, each given as the solution of a two-dimensional ODE system. Thus, each simulated individual belongs to one of two groups with distinct development patterns. 
We then simulate observations of $n \in \lbrace 100, 200 \rbrace$ individuals as follows: 
We randomly assign to each individual one of the two distinctly parametrized ODE systems as true underlying trajectories and sample an individual-specific second time point $t_{1,i}$ uniformly from the interval $[1.5,10]$. For both the common baseline timepoint $t_0 = 0$ and the second time point $t_{1,i}$, we then generate observations of $p=10$ variables. We sample five values from the first component of the individual's true trajectory at $t_0$ and $t_{1,i}$ for simulated variables $1$ to $5$, and five values from the second component for simulated variables $6$ to $10$, thereby reflecting the assumption of groups of variables sharing common trends. 
More specifically, for each time point we draw a variable-specific measurement error $\delta_{j}^{t_0}, \delta_{j}^{t_{1,i}} \sim \mathcal{N}(0,\sigma_{\mathrm{var}}^2)$ for $j=1,\dots, p=10$ and an individual-specific measurement error $\varepsilon_{i,j}^{t_0}, \varepsilon_{i,j}^{t_{1,i}} \sim \mathcal{N}(0,\sigma_{\mathrm{ind}}^2)$ for $i=1,\dots, n$, $j=1,\dots, p$ and add them to the true value of the corresponding component of the ODE solution at $t_0$ and $t_{1,i}$, respectively. Here, the standard deviations $\sigma_{\mathrm{var}}$ and $\sigma_{\mathrm{ind}}$ of the measurement error distributions determine the level of noise in the data. 

The values of additional baseline variables are either obtained by adding random noise to the individual's true ODE parameters, or by encoding the group membership of the individual as $+1$ or $-1$ and adding noise to these values. In the second variant, we thus only assume baseline variables to provide noisy information about the presence of two distinct groups in the dataset, making this a more realistic but also more challenging scenario. We draw a number of informative baseline variables from a normal distribution with mean corresponding to the group membership or ODE parameters, and standard deviation $\sigma_{\mathrm{info}}$. 
Additionally, we draw variables containing purely noise from a centered normal distribution with standard deviation $\sigma_{\mathrm{noise}}$, such that overall, we simulate a total of $q = 50$ baseline variables. 

\subsection{Linear ODE system with two unknown parameters}
\label{subsec:sim-linear2parameters}

The first simulation is based on two distinctly parametrized two-dimensional linear ODE systems, where the two non-zero parameters are assumed to be unknown:

\begin{minipage}{.4\linewidth}
	\begin{equation*}
	\begin{split}
	\frac{d}{dt}\begin{pmatrix} u_1 \\ u_2 \end{pmatrix}(t) &= \begin{pmatrix} -0.2 & 0 \\ 0 & 0.2 \end{pmatrix} \begin{pmatrix} u_1 \\ u_2 \end{pmatrix}(t); \\
	\begin{pmatrix} u_1 \\ u_2 \end{pmatrix}(0) &= \begin{pmatrix} 2 \\ 1 \end{pmatrix}
	\end{split}
	\end{equation*}
\end{minipage}\begin{minipage}{.5\linewidth}
	\begin{equation*}
	\begin{split}
	\frac{d}{dt}\begin{pmatrix} u_1 \\ u_2 \end{pmatrix}(t) &= \begin{pmatrix} -0.2 & 0 \\ 0 & -0.2 \end{pmatrix} \begin{pmatrix} u_1 \\ u_2 \end{pmatrix}(t); \\
	\begin{pmatrix} u_1 \\ u_2 \end{pmatrix}(0) &= \begin{pmatrix} 2 \\ 1 \end{pmatrix}
	\end{split}
	\end{equation*}
\end{minipage}

We simulate observations for $100$ individuals, $50$ from each of the two groups with a low level of noise ($\sigma_{\mathrm{var}} = \sigma_{\mathrm{ind}} = 0.1$), where $10$ baseline variables are simulated based on group membership, and $40$ noise baseline variables are added with $\sigma_{\mathrm{info}} = \sigma_{\mathrm{noise}} = 0.5$.

\begin{figure}
	\centering
	\begin{minipage}{\linewidth}
		\begin{minipage}{.5\textwidth}
			\centering				
			\includegraphics[width=\linewidth]{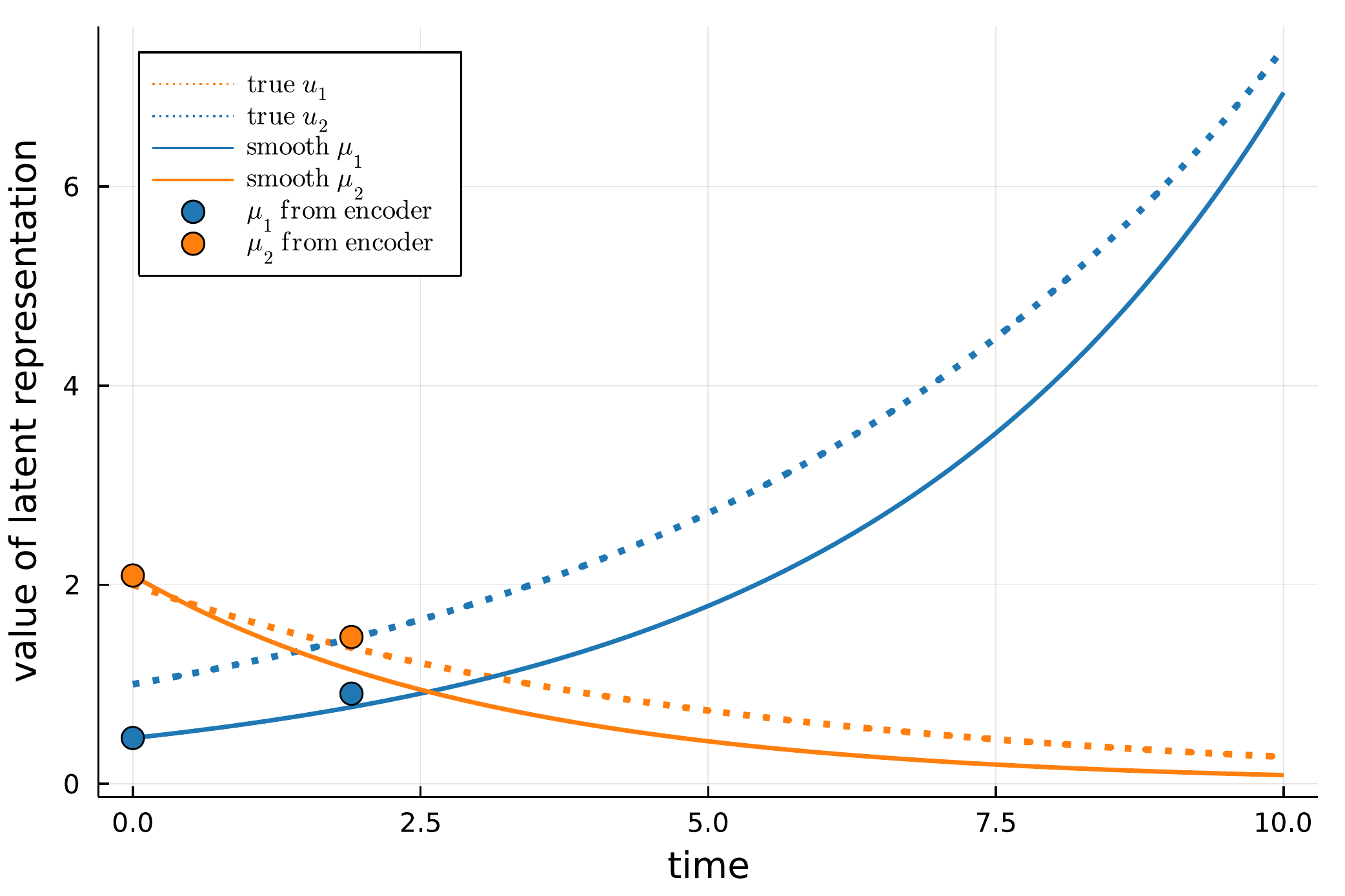}
		\end{minipage}\begin{minipage}{.5\textwidth}
			\centering
			\includegraphics[width=\linewidth]{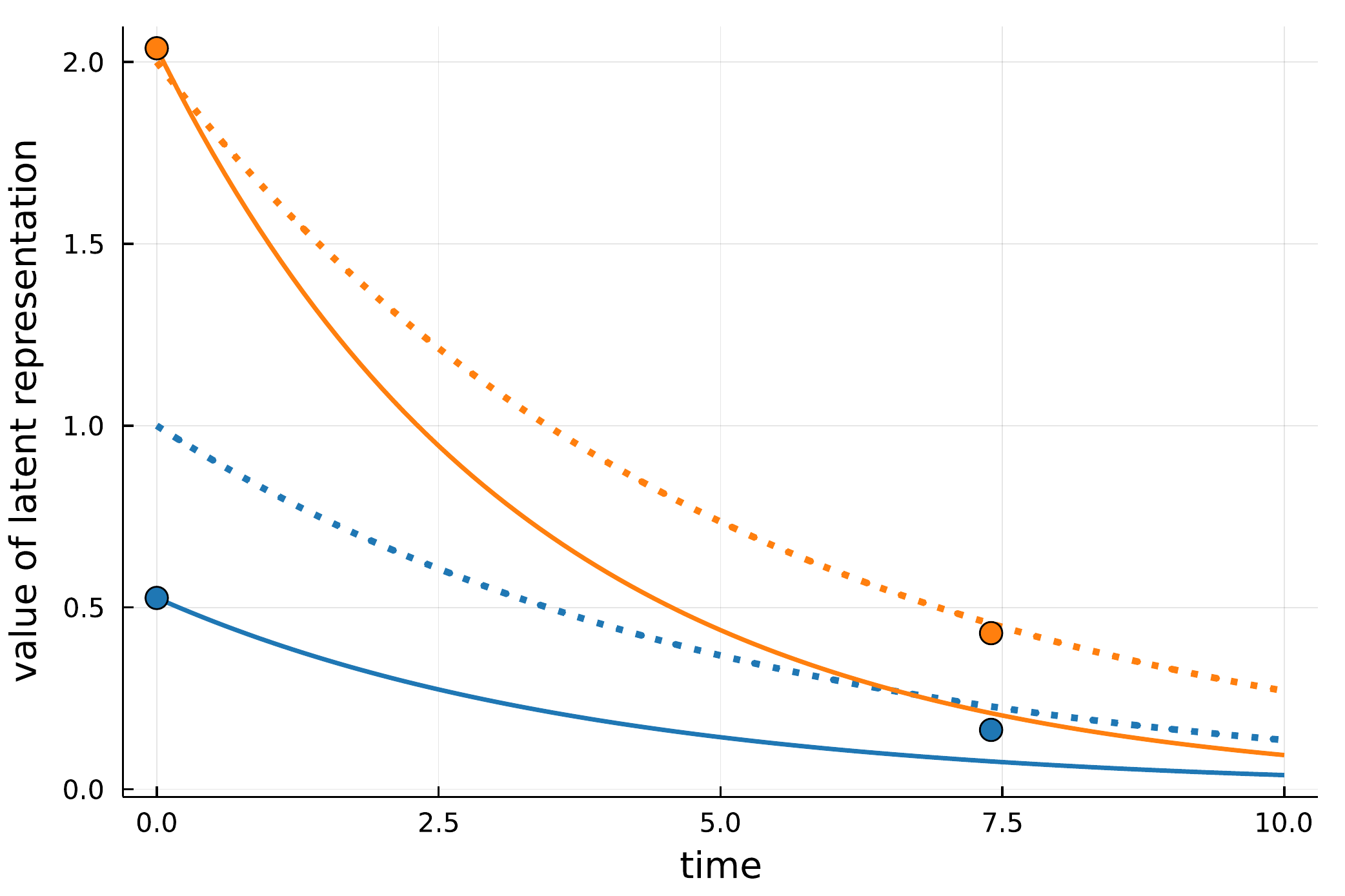}
		\end{minipage}
	\end{minipage}
	\begin{minipage}{\linewidth}
		\begin{minipage}{.5\textwidth}
			\centering
			\includegraphics[width=\linewidth]{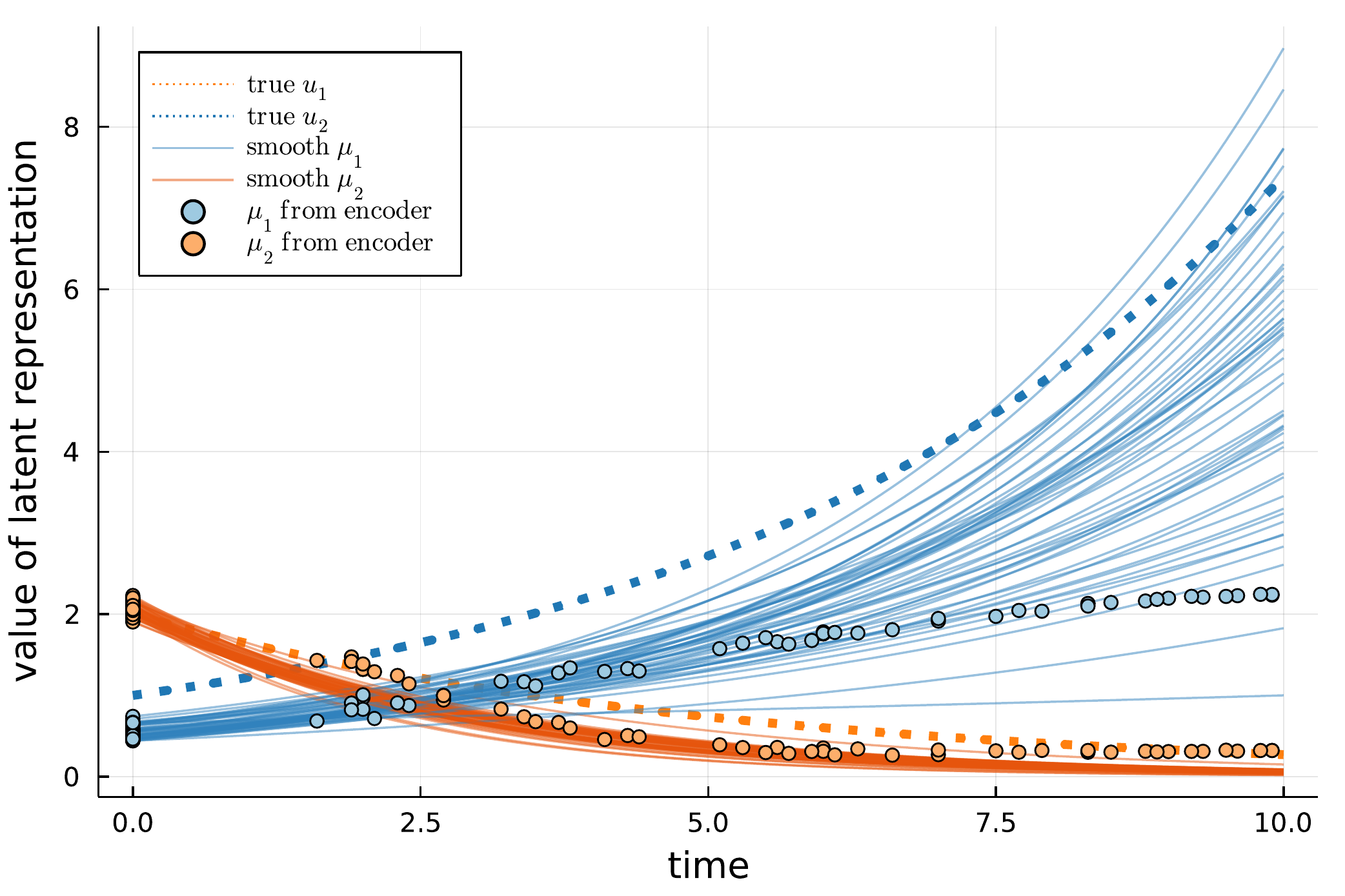}
		\end{minipage}\begin{minipage}{.5\textwidth}
			\centering
			\includegraphics[width=\linewidth]{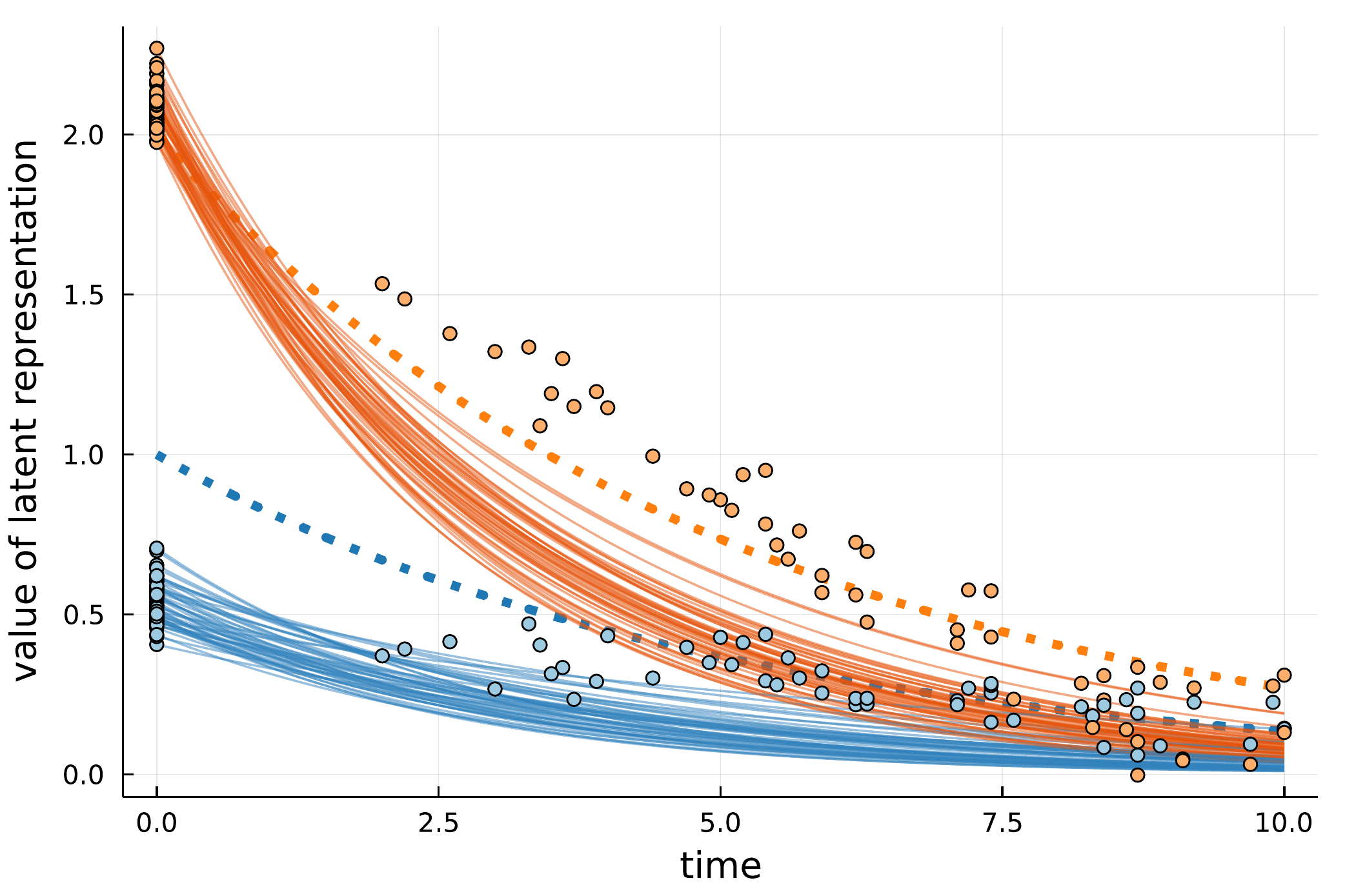}
		\end{minipage}
	\end{minipage}
	\caption{Latent representations and true ODE solutions of two individuals, one of each group (top row) and all individuals of each group (bottom row) for a linear ODE system.}
	\label{fig:singleinds_allinds_linear2_gi}
\end{figure}

In the first row of Figure~\ref{fig:singleinds_allinds_linear2_gi}, we show the resulting fitted ODE solutions of two individuals from each of the two groups. Here, the model recovers the general structure of both underlying patterns, recognizing the distinct upward/downward trends of variables for the two individuals. 
It tends to overestimate the initial value of the first component of the solution and to underestimate it in the second component. In both dimensions, the mean obtained directly from the VAE encoder before solving the ODE (dots) is consistent with the smooth mean obtained as ODE solution (solid lines), showing that in general the training strategy of matching the means before and after solving the ODE has the desired effect. 

The second row of Figure~\ref{fig:singleinds_allinds_linear2_gi} shows the ODE solutions for all individuals. While the model captures the exponential growth or decay of the fitted ODE solutions, the latent representation means obtained directly from the encoder before solving the ODE follow a linear rather than an exponential trend (blue dots in the bottom left and orange dots in the bottom right figure).

\subsection{Non-linear ODE system with two unknown parameters}
\label{subsec:sim-nonlinear}

\begin{figure}
	\centering
	\begin{minipage}{\linewidth}
		\begin{minipage}{.5\textwidth}
			\centering				
			\includegraphics[width=\linewidth]{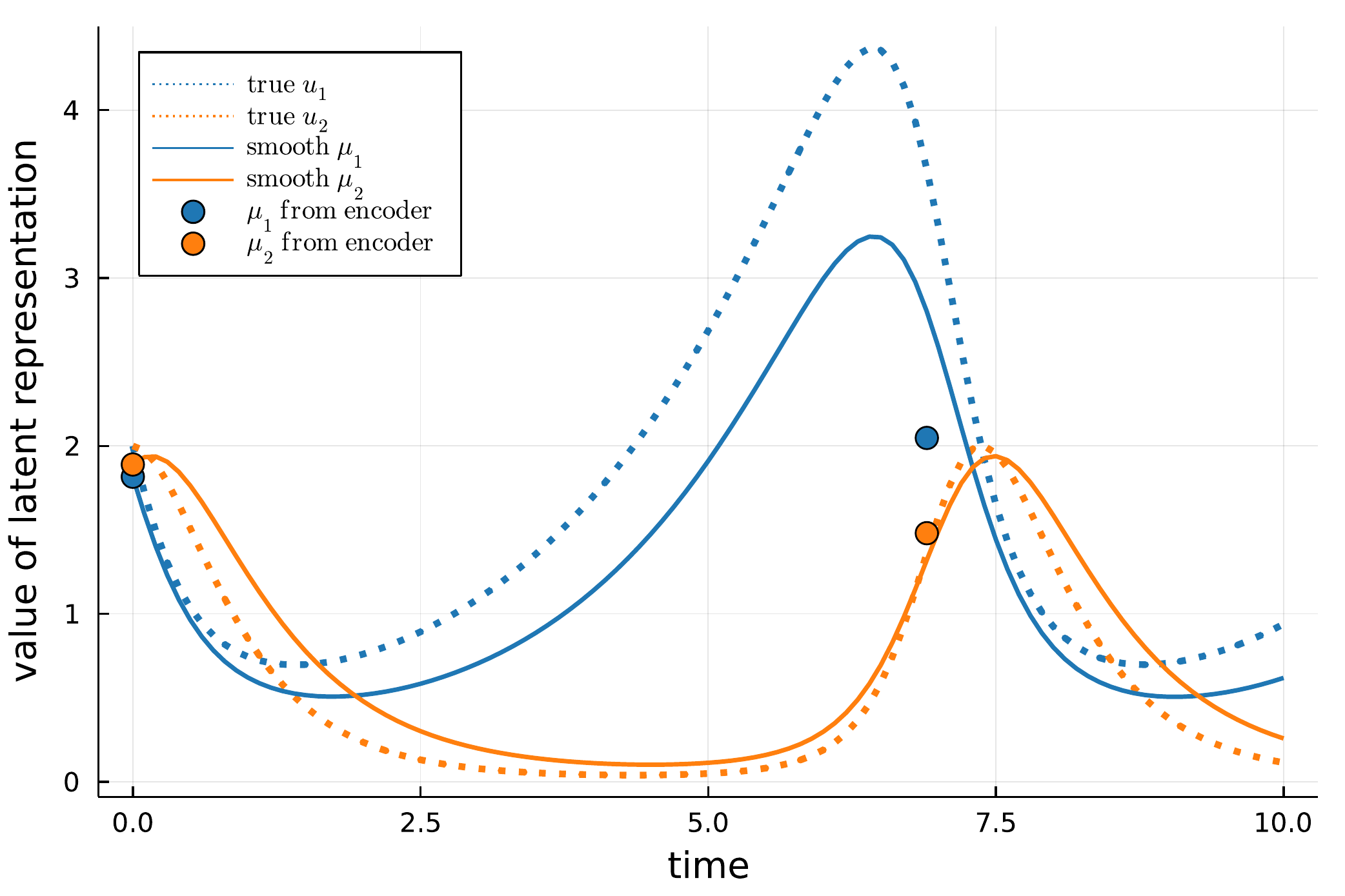}
		\end{minipage}\begin{minipage}{.5\textwidth}
			\centering
			\includegraphics[width=\linewidth]{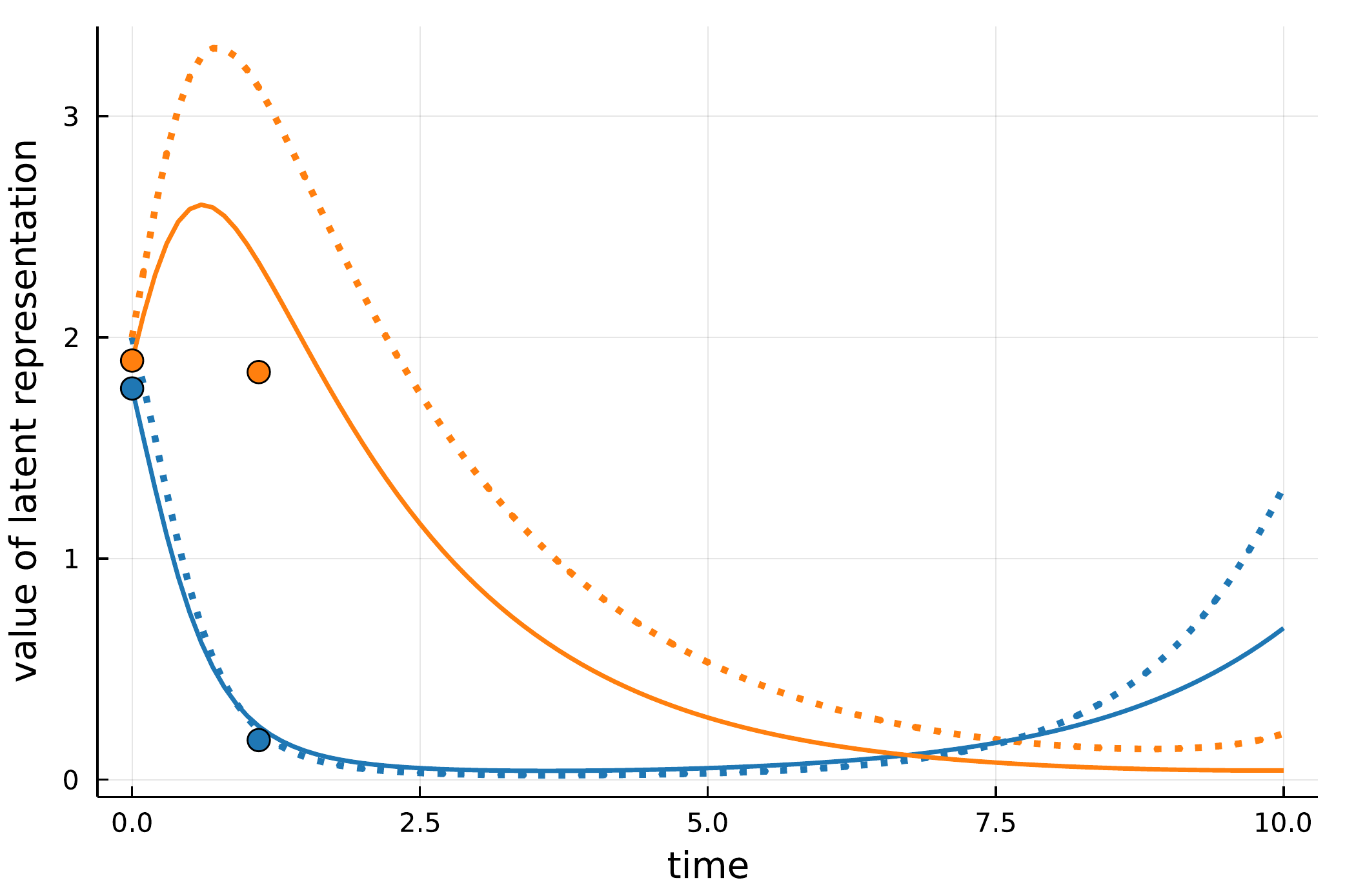}
		\end{minipage}
	\end{minipage}
	\begin{minipage}{\linewidth}
		\begin{minipage}{.5\textwidth}
			\centering
			\includegraphics[width=\linewidth]{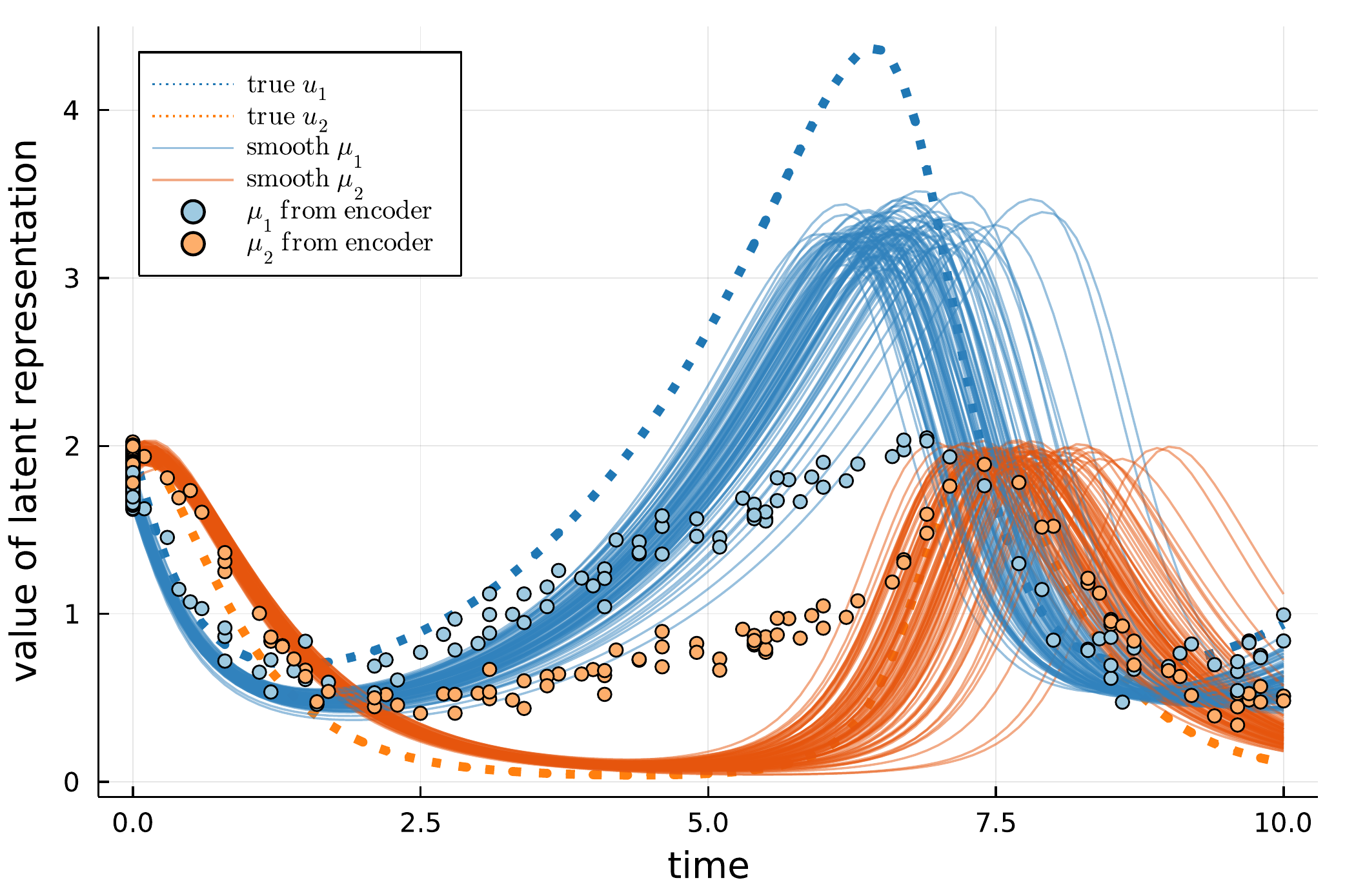}
		\end{minipage}\begin{minipage}{.5\textwidth}
			\centering
			\includegraphics[width=\linewidth]{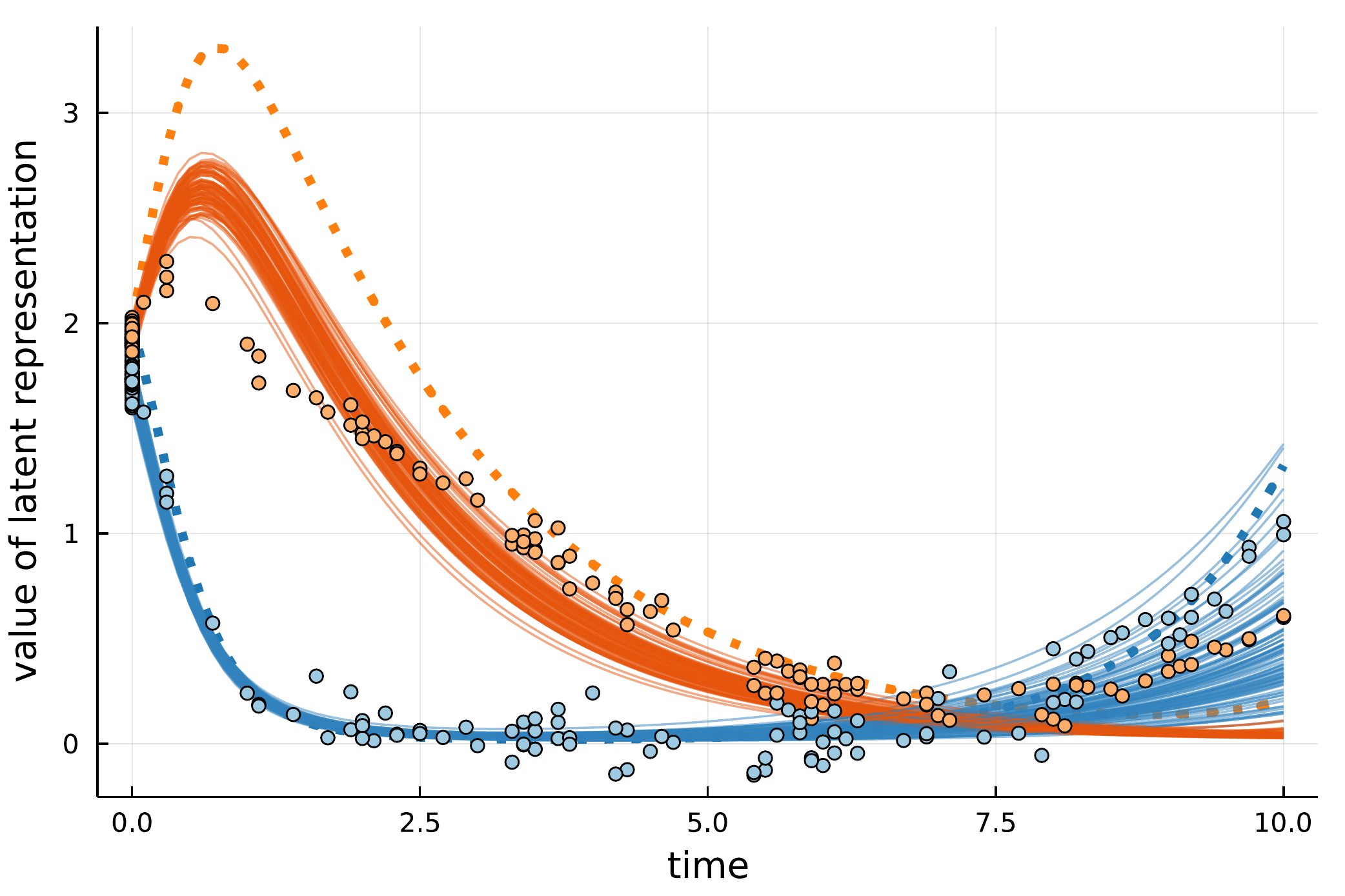}
		\end{minipage}
	\end{minipage}
	\caption{Latent representations and true ODE solutions of two individuals, one of each group (top row) and all individuals of each group (bottom row) for a non-linear ODE system.}
	\label{fig:singleinds_allinds_lv_gi}
\end{figure}

To investigate whether our method can also capture non-linear underlying developments, we simulate data from two distinctly parametrized non-linear Lotka-Volterra systems:

\begin{minipage}{.4\linewidth}
	\begin{equation*}
	\begin{split}
	\frac{d}{dt}\begin{pmatrix} u_1 \\ u_2 \end{pmatrix}(t) &= 
	\begin{pmatrix} 0.5 u_1(t) - u_1(t) u_2(t) \\ u_1(t) u_2(t) - 2 u_1(t) \end{pmatrix}; \\
	\begin{pmatrix} u_1 \\ u_2 \end{pmatrix}(0) &= \begin{pmatrix} 2 \\ 2 \end{pmatrix}
	\end{split}
	\end{equation*}
\end{minipage}\begin{minipage}{.55\linewidth}
	\begin{equation*}
	\begin{split}
	\frac{d}{dt}\begin{pmatrix} u_1 \\ u_2 \end{pmatrix}(t) &= 
	\begin{pmatrix} u_1(t) - u_1(t) u_2(t) \\ u_1(t) u_2(t) - 0.5 u_1(t) \end{pmatrix}; \\
	\begin{pmatrix} u_1 \\ u_2 \end{pmatrix}(0) &= \begin{pmatrix} 2 \\ 2 \end{pmatrix}	
	\end{split}
	\end{equation*}
\end{minipage}

Here, we generated observations of $200$ individuals, $100$ from each group, to account for the higher complexity of the true system and to more densely cover the entire trajectory with the randomly drawn second measurement time points. Again, we set $\sigma_{\mathrm{var}} = \sigma_{\mathrm{ind}} = 0.1$ and simulate baseline variables drawing from the group membership encoded as $\pm1$, using $30$ informative baseline variables and adding $20$ noise variables with $\sigma_{\mathrm{info}} = \sigma_{\mathrm{noise}} = 0.5$. 

Again, we show both individual solutions and the solutions across all individuals from the two groups (Figure~\ref{fig:singleinds_allinds_lv_gi}). Similar to the linear setting, in this scenario the model also underestimates the high peaks of the first component $u_1$ for the first underlying pattern and of the second component $u_2$ in the second underlying pattern, in particular with respect to the means of the latent representation obtained directly from the encoder without solving the ODE. This could be due to the regularizing effect of the $\mathcal{N}(0,1)$-prior on the latent space, which prevents extreme values. Parallel to the linear setting, this phenomenon is less pronounced for fitted ODE solutions which can be influenced by the baseline information. While here the model also underestimates the peak in the first component of the pattern of the first group (blue lines in left panels), it roughly learns the underlying pattern in both groups. In this setting, we additionally investigated the effect of different numbers of informative baseline information on the model performance and found that with lower numbers, the model cannot clearly distinguish between the distinct underlying patterns (results not shown). 

\subsection{Linear ODE system with four unknown parameters}
\label{subsec:sim-linear4parameters}

Finally, we investigate whether the approach described in Section~\ref{subsec:trainingonbatches} allows for reliably estimating four unknown ODE parameters from baseline variables, i.e., more parameters than time points available. For this, we define two distinct development patterns based on a linear ODE system as follows:

\begin{minipage}{.4\linewidth}
	\begin{equation*}
	\begin{split}
	\frac{d}{dt}\begin{pmatrix} u_1 \\ u_2 \end{pmatrix}(t) &= \begin{pmatrix} -0.2 & 0.1 \\ -0.1 & 0.25 \end{pmatrix} \begin{pmatrix} u_1 \\ u_2 \end{pmatrix}(t); \\
	\begin{pmatrix} u_1 \\ u_2 \end{pmatrix}(0) &= \begin{pmatrix} 4 \\ 2 \end{pmatrix}
	\end{split}
	\end{equation*}
\end{minipage}\begin{minipage}{.55\linewidth}
	\begin{equation*}
	\begin{split}
	\frac{d}{dt}\begin{pmatrix} u_1 \\ u_2 \end{pmatrix}(t) &= \begin{pmatrix} -0.2 & 0.1 \\ 0.1 & -0.2 \end{pmatrix} \begin{pmatrix} u_1 \\ u_2 \end{pmatrix}(t); \\
	\begin{pmatrix} u_1 \\ u_2 \end{pmatrix}(0) &= \begin{pmatrix} 4 \\ 2 \end{pmatrix}
	\end{split}
	\end{equation*}
\end{minipage}

We generate observations of $10$ variables for $100$ individuals, $50$ from each of the two groups, as before, employing a medium level of noise with $\sigma_{\mathrm{var}}= 0.1$ and $\sigma_{\mathrm{ind}} = 0.5$. In this substantially more difficult setting, we present results based on using the true ODE parameters as baseline information, where for each parameter, we simulate $5$ baseline variables by sampling from that parameter and add $30$ purely noise variables, setting $\sigma_{\mathrm{info}} = \sigma_{\mathrm{noise}} = 0.1$. We train the model on groups of similar individuals as described in Section~\ref{subsec:trainingonbatches} and now estimate all four parameters as outputs of the ODE-net. 

Regarding the batch size, we found empirically that a small batchsize ($b \leq 5$) is insufficient to capture the complete dynamics because it provides too little time points and proxy observations. On the other hand, a large batchsize with respect to the dataset size ($b \gg \frac{n}{10}$) introduces training instability due to too many individuals with different development pattern being grouped together. Between those extremes, we found the results to be robust against different choices of batchsizes and present in the following results for a batchsize of $b=10$ only. We also found the results to be stable with respect to different choices of kernel functions and report results for a tricube kernel of bandwidth $1$. 

\begin{figure}
	\centering
	\begin{minipage}{\linewidth}
		\begin{minipage}{.5\textwidth}
			\centering				
			\includegraphics[width=\linewidth]{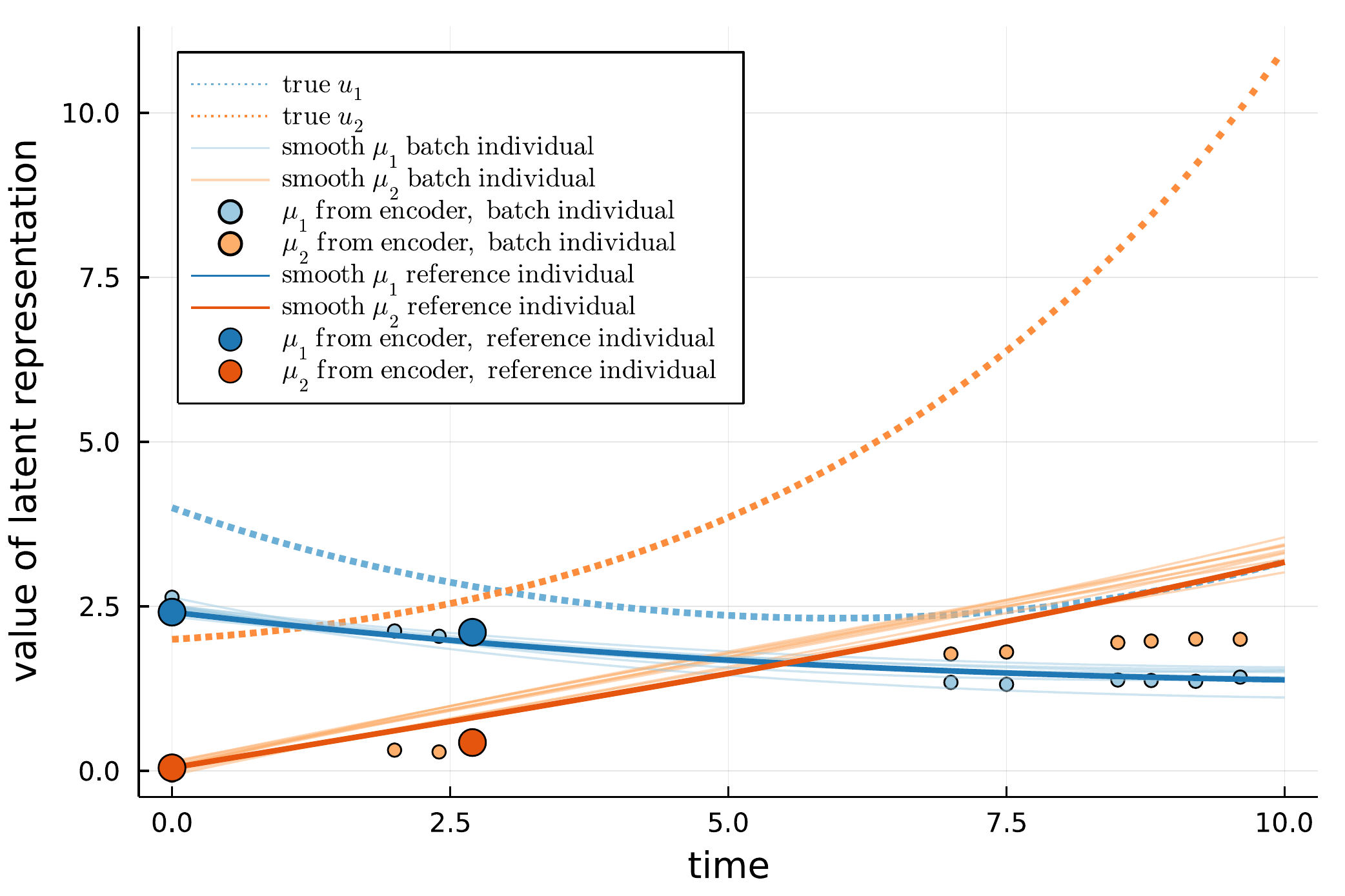}
		\end{minipage}\begin{minipage}{.5\textwidth}
			\centering
			\includegraphics[width=\linewidth]{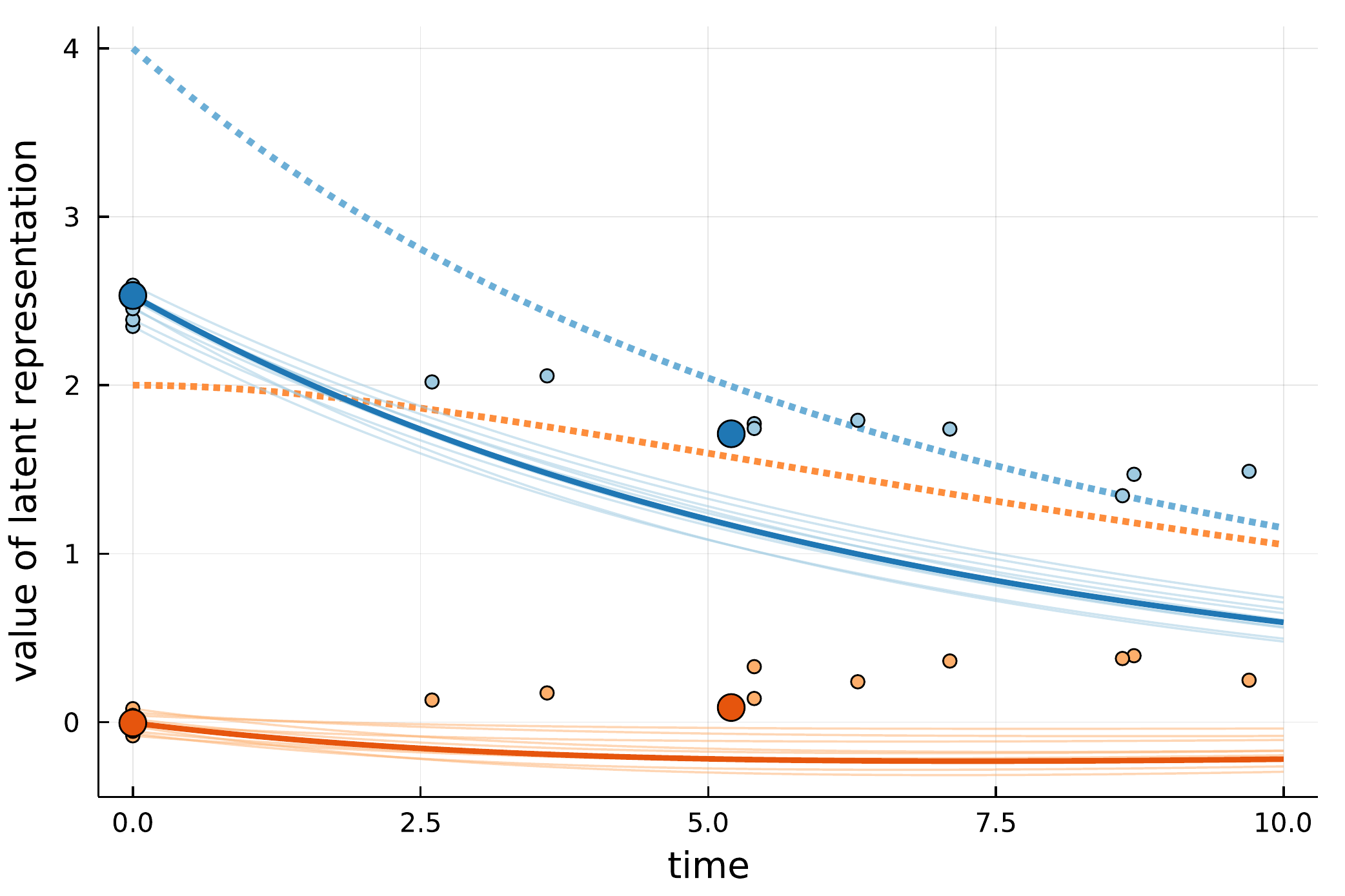}
		\end{minipage}
	\end{minipage}
	\begin{minipage}{\linewidth}
		\begin{minipage}{.5\textwidth}
			\centering
			\includegraphics[width=\linewidth]{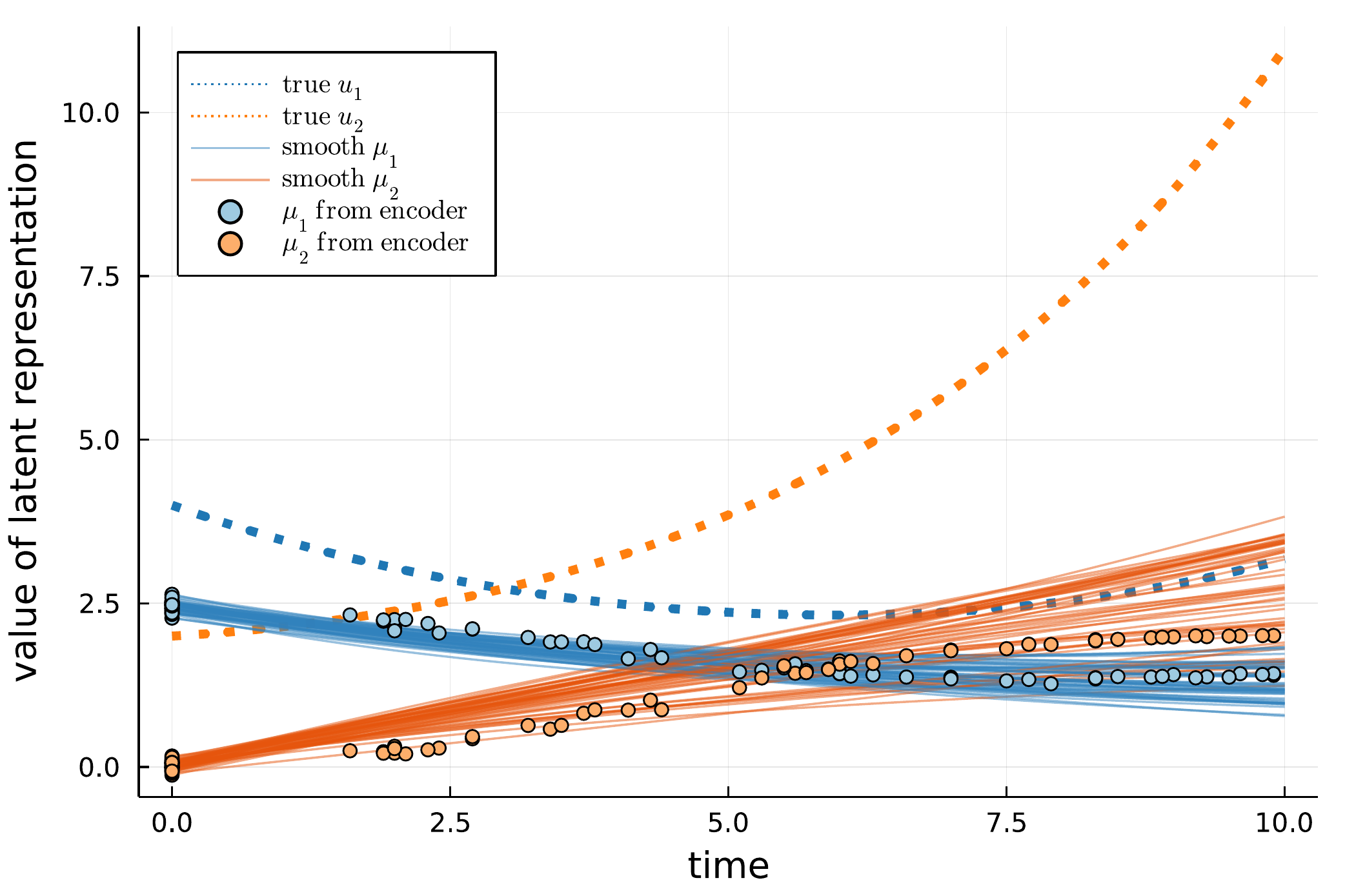}
		\end{minipage}\begin{minipage}{.5\textwidth}
			\centering
			\includegraphics[width=\linewidth]{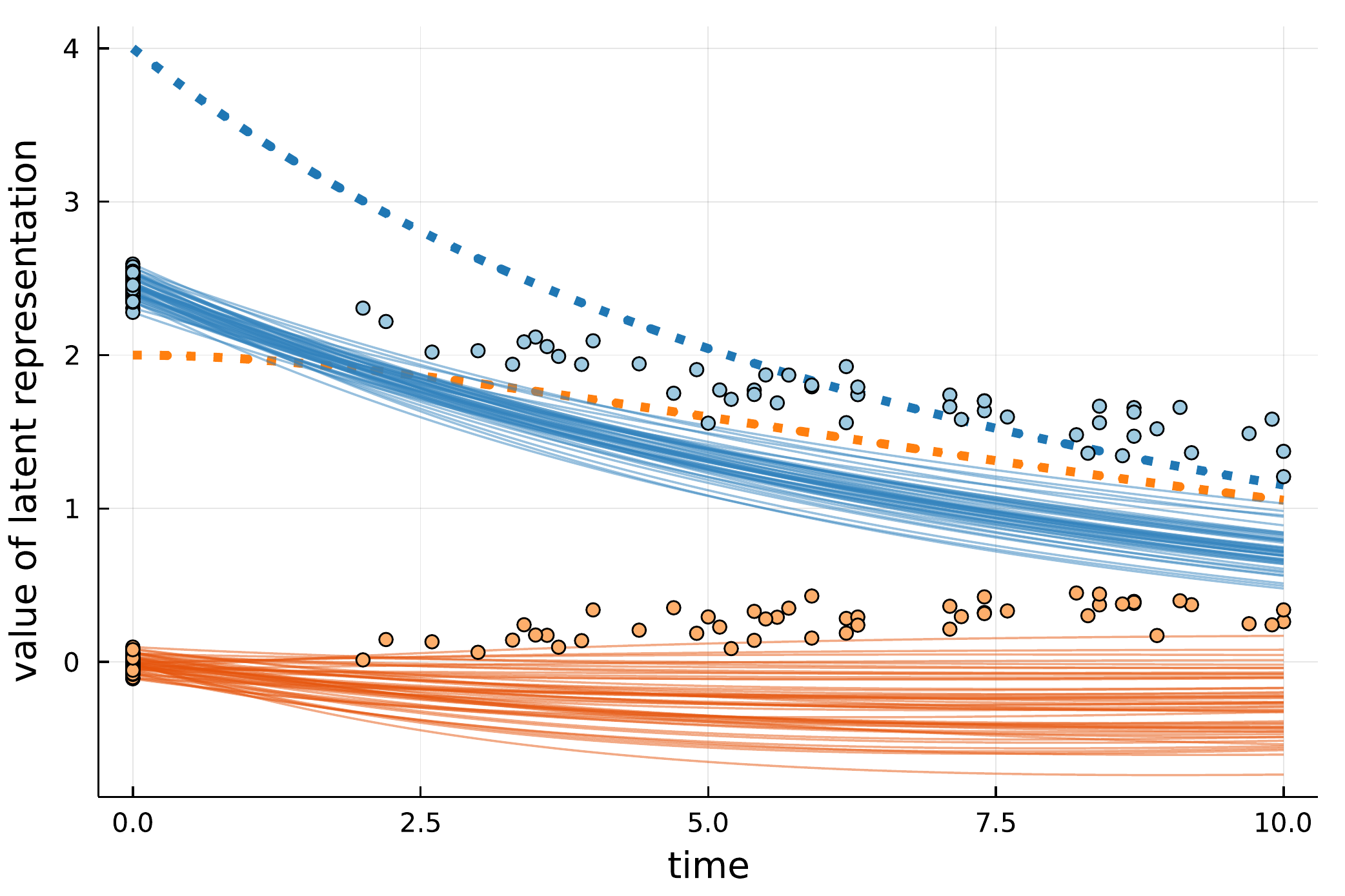}
		\end{minipage}
	\end{minipage}
	\caption{Latent representations and true ODE solutions of one selected batch of each group (top row) and of all individuals from each group (bottom row) In the top row, one panel depicts the fitted ODE solutions of all individuals in the batch. Thick, dark lines and dots represent the latent representation means of the reference individual around which the batch was grouped, while lighter, thin lines and dots belong to the other individuals in the batch.}
	\label{fig:singlebatch_allinds_linear4_tp}
\end{figure}

In Figure~\ref{fig:singlebatch_allinds_linear4_tp}, we show the ODE solutions of individual batches for both groups of underlying ODE systems in the first row and the solutions of all individuals in each group in the second row.	
With respect to the individual batches in the first row, we observe that all individuals in the group display similar ODE solutions and have thus been assigned similar ODE parameters. The second measurement time points of all individuals in the batch cover the entire time interval. While the ODE solutions of all individuals in the group are reflecting the correct trends of the true development pattern, they underestimate the true solution, especially in the first component. This is even more pronounced in the first component of the means obtained from the VAE encoder before solving the ODE. However, these means also reflect the true developments with some variability (the apparently higher variability in the figures on the right is due to the different scaling of the $y$-axis).
Again, particularly the upward trend of the second component of the ground-truth trajectory is underestimated in the means from the encoder, while the fitted ODE solutions show that the integrated ODE solving step can partially improve this. 
The variability of the means from the encoder tends to be higher later in the time interval, reflecting the greater uncertainty in the fitted ODE solutions at those later time points. While for the shown scenario, the general trends of the trajectories are still distinguishable, when training with the group membership as baseline information, they tend to become superimposed by the increased variability between individual's solutions (results not shown). Additionally, here we frequently observe that components of the fitted solution are flipped at the $x$-axis. We conclude that to estimate more parameters than time points, the model has to rely more heavily on the information in the baseline variables to still identify the individual dynamics, and the group membership alone is not sufficient. 

We also found that when training the model without grouping individuals based on similarity, it is not able to identify the two distinct development patterns at all, and that the performance is significantly worse if random batches are used. This shows that the model does in fact profit from taking into account individuals' similarity. 

\section{Exemplary application on data from the SMArtCARE registry}\label{sec:SMArtCAREapplication}

To complement the simulation study results, we additionally applied the approach to data from the SMArtCARE registry, a prospective multicenter cohort study of spinal muscular atrophy (SMA) patients that collects longitudinal data on patients' disease developments during routine visits \citep{Pechmann2019}.

While the registry is the world's largest data collection on SMA patients, the application still constitutes a small data setting from a modeling perspective, as outlined in the introduction. Specifically, as the registry has been set up only recently, there are not many time points per patient available yet, with considerable variability in timing and frequency of follow-up visits. Also, the patient population is more heterogeneous compared to previous clinical studies. Correspondingly, we investigate to what extent our modeling approach can extract individual-specific disease trajectories to account for heterogeneity in this challenging setting.

In addition to an extensive baseline characterization, the motor function of every SMArtCARE patient is assessed regularly at routine visits with one or more different physiotherapeutic assessments selected according to the severity of symptoms, the motoric ability and age of the patient. Additionally, during every patient visit, data on respiratory and nutritional status (e.g., does the patient require ventilation or a feeding tube), experience of pain, fatigue or adverse events such as hospitalization are recorded, among others. For a more detailed description of data collection and a complete overview of time-dependent and baseline variables, please see \citet{Pechmann2019}. 

For our application, we selected patients treated with Nusinersen, the first drug approved for the treatment of all SMA patients \citep{Schorling2020}, who are typically seen at regular 4-months intervals, but time intervals are sometimes longer if routine visits are missed or postponed. To have a common set of variables for all patients considered in our analysis, we selected a specific motor function test conducted for broader range of patients, the Revised Upper Limb Module (RULM) \citep{Mazzone2017} that evaluates motor function of upper limbs and can be conducted for all patients older than two years of age with the ability to sit in a wheelchair \citep{Pechmann2019}. 
We used all individual items and the overall score of the test as well as additional time-dependent information on respiratory and nutritional status, pain, fatigue and adverse events, and all available baseline characteristics including, e.g., sex, age at symptom onset, and genetic test results.
Overall, our data set comprises observations of $32$ time-dependent variables and $24$ baseline variables in $503$ individuals at between $2$ and $12$ time points (median: $5$ time points), resulting in a total of $3176$ single observations. The reference time point for all individuals is given by the time point of the first baseline visit, i.e., it is predetermined by the application and fixed at $t_0$. 

On this data, we applied our modeling approach as described in Section \ref{sec:methods}, using a two-dimensional latent space linear ODE system with two unknown parameters as in the simulation scenario of Section \ref{subsec:sim-linear2parameters}, i.e., allowing no interaction between the two latent space dimensions. We trained the model in exactly the same framework as for the simulation design, without specifically tuning the approach or its hyperparameters, other than choosing a learning rate and number of training epochs based on monitoring convergence of the loss function. We thus aimed to keep the model and optimization as simple as possible, to limit potential overfitting and give a realistic impression of the method's capacity and limits when applying it naively without extensive fine-tuning. 
This also applies to the choice of the underlying ODE system where in principle, clinical domain knowledge could be used, but we deliberately chose to keep it simple. 

\begin{figure}
	\includegraphics[width=\linewidth]{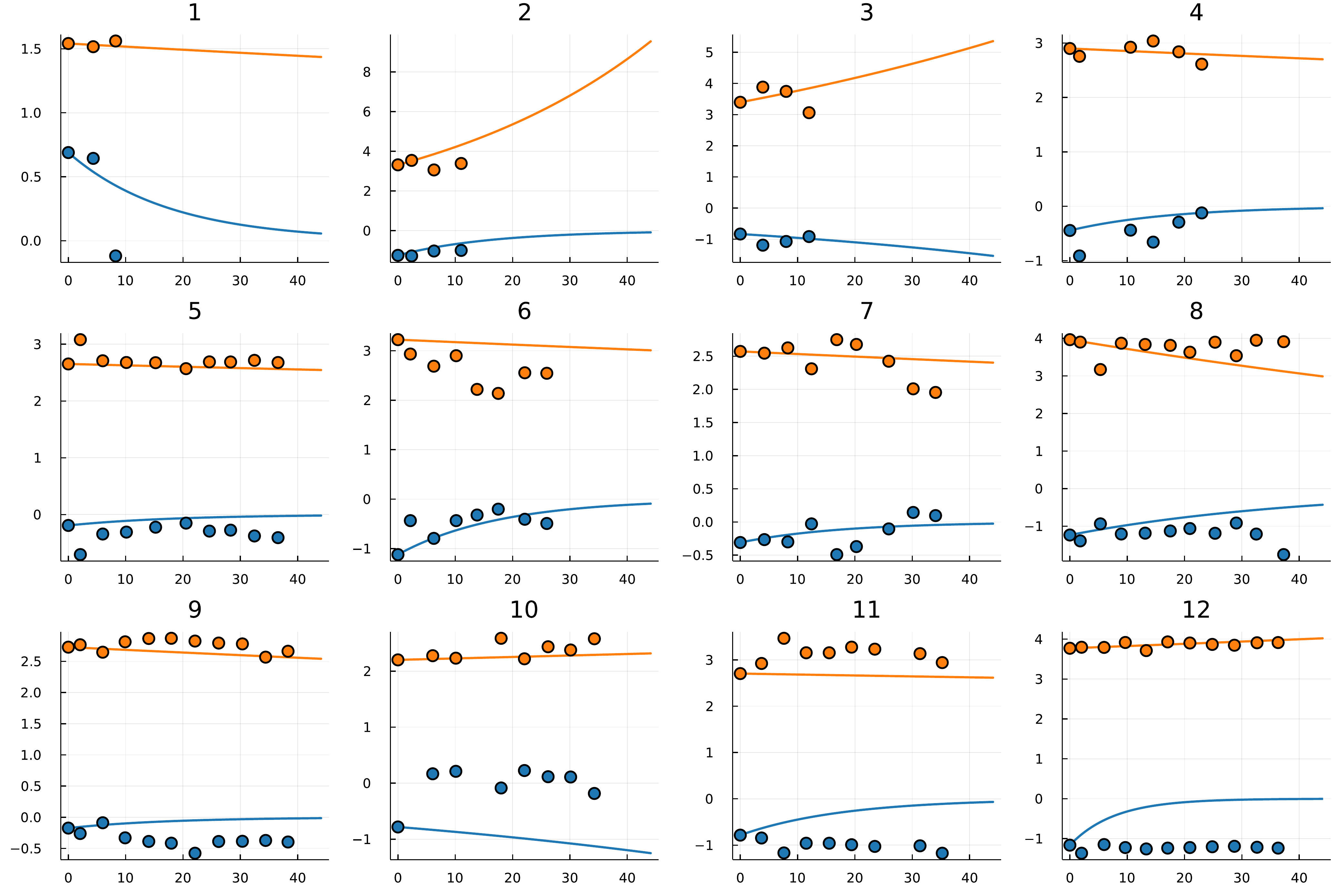}
	\caption{Exemplary learnt latent representations from data of SMArtCARE patients. Each panel shows the learnt latent trajectories of one patient. The $x$ axis shows the time in months and the $y$ axis the value of the latent representation. Dots depict the means of the encoder obtained directly from applying the VAE encoder to the observations at each time point, solid lines depict the fitted ODE solutions. Blue dots and lines correspond the first latent dimension and orange dots and lines to the second dimension.}
	\label{fig:SMArtCAREexamples}
\end{figure}

In Figure \ref{fig:SMArtCAREexamples}, we show learnt latent trajectories from $12$ exemplary SMArtCARE patients. Note that visual inspection and interpretation of learnt patterns is complicated because the model is free to arbitrarily scale and shift its representation and, e.g., flip it at the $x$ axis, which makes it difficult to compare representations across different models fits, e.g., to assess model variability, leading us to show only exemplary fits. 

The number of available time points varies, and while time intervals are often regular, there are also longer stretches with no observation (e.g., for the patient in panel 4, there is a larger gap between the second and third measurement). Generally, we can observe that for the patients in the two upper rows, the model captures underlying upward or downward trends and the encoded data (i.e., the means from the encoder corresponding to the blue and orange dots) match the fitted trajectories (blue and orange lines). 
In the third row, we show representations of patients where the encoded data does not match the fitted trajectories in at least one of the components. We chose a representative selection of examples for visualization in accordance with our general aim to investigate both the capacity and breaking point of our modeling approach.
Further, it can be seen that for some patients, these trajectories exhibit a distinct upward or downward trend, such as for the blue component in panel $1$ and the orange component in panel $2$, while for other patients the model fits rather constant trajectories (e.g., panel $5$). The modeling approach can thus capture individual-specific dynamics that reflect the heterogeneity of the patient population. 

\section{Discussion}
\label{sec:discussion}

While longitudinal biomedical data from individuals promises insights into the development of an underlying latent health status, extracting these trajectories on an individual level is often impaired by a small number of observed time points. For example when setting up epidemiological cohorts or clinical registries, initially only two time points, at baseline and one follow-up measurement, might be available. 
Deep learning techniques, which have predominantly been applied on large datasets, have recently also been shown to uncover complex patterns in data with small sample sizes \citep{Hess2017, Nussberger2020}. 
Yet, such approaches are often viewed as black-box algorithms that lack interpretability and a thorough theoretical underpinning. This is partly addressed by recent developments on combining deep learning with explicit modeling \citep{Chen2018, Rackauckas2020} and establishing the field of scientific machine learning \citep{Innes2019_Zygote_dP}. These techniques also promise performance improvements with a limited amount of information. Therefore, we set out to explore whether such approaches might also be useful for the described biomedical setting with just two time points, and where they break down. 
Identifying these breakdown points can then guide a more informed decision on whether a complex model is feasible or a simpler modeling approach would be better suited for the scenario at hand. It can thus inform further model selection and also help to prevent potential overfitting with an overly complex model.

Specifically, we developed an approach that incorporates structural assumptions in a generative deep learning framework and explored its feasibility to infer individual-specific trajectories. 
The approach encodes observations from both time points in a latent space defined by a random variable. Here, we solve an ODE system for the mean of the latent variable and adapt the training objective to the extremely sparse time grid by explicitly encouraging consistency of the latent representation means before and after solving the ODE. By describing the latent space by smooth trajectories obtained as ODE solutions, we effectively constrain the latent representation to model smooth dynamics. Imposing the ODE system thus has a regularizing effect by enforcing smoothness of the latent representation. With a simple choice of ODE system, this can thus counteract potential overfitting due to the limited amount of information. 
Further, we focused on how such an approach could additionally benefit from using implicit regularity assumptions in the mapping of baseline variables to dynamic model parameters. Furthermore, we investigated an iterative optimization framework based on training the model on groups of similar individuals. 
We also described how our approach fits into a filtering framework, for a statistical perspective. While this suggests that the approach encompasses classical approaches such as Kalman filters as special case, a more detailed comparison is beyond the scope of the present paper. Nonetheless, it would be promising to investigate in future research whether also other approaches could be adapted for being feasible with just two time points, and how the underlying theory relates to our proposed approach. 

We simulated data based on the described scenario by defining the true underlying ODE systems and sampling from the trajectories. For both a linear and a non-linear ODE system with two unknown parameters, the proposed approach could recover the distinct underlying temporal development patterns and infer individual-specific ODE parameters. To identify settings where such a combined neural network and dynamic modeling approach might break down, we also considered a setting with four unknown ODE parameters. In this challenging setting, the approach could still identify groups of individuals with similar trajectories and infer basic underlying trends, but requires considerably more information from the additional baseline variables.

To explore the feasibility of the approach on real clinical data, we additionally applied the approach on a data set from the SMArtCARE registry characterized by a varying, but generally small number of time points per patient, without further fine-tuning or specific adaptation to the data set. The results show that the modeling approach can still infer individual-specific latent trajectories for most patients, but not in all cases. Thus, this provides an exemplary application at the verge of what the model is capable of fitting when applied naively without further adjustment to the specific setting. 
Yet, such tuning and adaptation will be needed for a clinical interpretation and meaningful implications for clinical investigators, which has yet to be addressed in future research. In particular, here the generative capacity of the model can be utilized to generate latent trajectories and decode them to synthetic observations in data space, that can subsequently be queried for clinically meaningful trends. 

In the present work, our focus was to explore whether the model can even be fit to such challenging data scenarios and extract individual-specific dynamics despite a small number of time points. Accordingly, our simulation investigated to what extent such an approach can handle the challenging task of inferring individual dynamics from only two time points rather than extensively testing model and data variations. We thus specifically optimized our model with respect to this small data setting. 

As a limitation, the dynamic model is currently deterministic and does not allow for measurement times that depend on the latent dynamic process. Therefore, an extension to stochastic differential equations is a logical next step, which would, e.g., allow to model variations in disease processes due to unexplained influences. 
Due to the underlying VAE architecture, the approach is not restricted in its embedding of the data in latent space. This might also mean that the occasionally occurring sign changes or swaps of dimension might preclude straightforward intuitive interpretation of the learnt latent representation. This limitation could be addressed in future work by explicitly encouraging the model to find a latent representation satisfying certain constraints. 
Additionally, the approach is sensitive to noisy information in the time-dependent and baseline variables as well as to how the data is scaled with respect to the $\mathcal{N}(0,1)$-prior. This becomes particularly apparent in the challenging scenario of identifying four unknown ODE parameters, such that in its current form, the approach is restricted to rather specific settings. 

In our present work, we focused on evaluating whether a deep learning-based method could potentially be useful for the task at hand, and to exemplarily investigate how integrating structural assumptions both in the form of dynamic modeling components and similarity and filtering in a deep learning approach could allow to identify individual temporal development patterns in a challenging small data scenario. We envision that further developing such approaches bears the potential to provide an individual-level understanding of the temporal dynamics underlying individuals' developments and, rather than estimating average effects, has the potential to plan interventions based on the knowledge of full individual-specific dynamical systems.

\vspace*{1pc}

\noindent {\bf{Acknowledgement}}

This work was supported by the DFG (German Research Foundation) -- 322977937/GRK2344 (MH). Biogen and Novartis provide financial support for the SMArtCARE registry. AP was supported by the Berta-Ottenstein clinician scientist program of the University of Freiburg.
\vspace*{1pc}

\noindent {\bf{Conflict of Interest}}

\noindent {\it{The authors have declared no conflict of interest.}}

\section*{Appendix}

\subsection*{A.1.\enspace Implementation details}

The following section provides details on the implementation of the proposed approach. 
Additionally, the complete code and Jupyter notebooks to reproduce our simulation study results and illustrate the usage of the model can be found at \url{https://github.com/maren-ha/DeepDynamicModelingWithJust2TimePoints}. 

The code to run all simulations and models is written in the {\tt{Julia}} programming language \citep{Bezanson2017} of version 1.6.3 with the additional packages {\tt{BenchmarkTools.jl}} (v.1.2.0), {\tt{CSV.jl}} (v0.9.10), {\tt{DataFrames.jl}} (v1.2.2), {\tt{DiffEqFlux.jl}} (v1.44.0), {\tt{Distributions.jl}} (v0.25.24), {\tt{Flux.jl}} (v0.12.8), {\tt{JLD2.jl}} (v0.4.15), {\tt{LaTeXStrings.jl}} (v1.3.0), {\tt{OrdinaryDiffEq.jl}} (v5.65.5), {\tt{Plots.jl}} (v1.23.5) and {\tt{VegaLite.jl}} (v2.6.0). 

For the VAE, we aimed at keeping the architecture simple and close to established architectures from the literature. Specifically, we use one hidden layer with the number of hidden units equal to the number of input dimensions (i.e., $10$ in all applications) and a $\tanh$-activation function shifted by $+1$. 
The latent space is two-dimensional, and the latent space mean and variance are obtained as affine linear transformations of the hidden layer values without a non-linear activation. The decoder includes one hidden layer with $10$ hidden units and a $\tanh$-activation function. It outputs the mean and variance of a Gaussian distribution which are calculated from the hidden layer using an affine linear transformation. 
In the loss function, the KL-divergence between the prior and posterior is scaled by a factor of $0.5$ to reduce the regularizing effect. The sum of squared values of all decoder parameters with a weighting factor of $0.01$ is added as commonly used penalty term to prevent exploding model parameters. 

The ODE-net consists of three layers in addition to the input layer, two hidden layers and one output layer. In the first hidden layer, the number of units equals the number of input dimensions (i.e., $50$ in all the applications) and a $\tanh$-activation function is used. In the second layer, the number of hidden units equals the number of parameters to be estimated and the activation function is a shifted sigmoid function. The shifting of the activation function serves as a prior to guide the model in identifying the range in which to estimate the ODE parameters: For the applications with linear underlying ODE systems and true parameters of around $0.2$ or $-0.2$ (Sections~\ref{subsec:sim-linear2parameters} and \ref{subsec:sim-linear4parameters}), the sigmoid function is shifted by $-0.5$ such that it outputs values in the interval $[-0.5,0.5]$. For the applications with non-linear underlying ODE systems and true parameters between $0.5$ and $2$ (Section~\ref{subsec:sim-nonlinear}), it is shifted by $+1$, such that it outputs values in $[0,2]$. The model is not restricted to these ranges because an affine linear transformation with a diagonal matrix is added as a final output layer. 

All models were trained with the ADAM SGD-optimizer \citep{Kingma2015} using learning rates between $0.001$ and $0.0001$ and between $15$ and $40$ training epochs. We chose the numbers of epochs and learning rates based on monitoring convergence of the loss function and visualizing training results. 

For the applications with the non-linear ODE system from Section~\ref{subsec:sim-nonlinear}, we frequently observed training runs where the random weight initialization of the network produced instabilities in the numerical solution of the ODE with the Tsit5-ODE solver implementing a Tsitouras 5/4 Runge-Kutta method. To reduce the number of resulting training break-offs, we changed the default initialization method for the weights to smaller values, which resulted in less solver instabilities but did not completely prevent them.

Generally, backpropagating through the ODE solving step within the neural network framework is realized by the differentiable programming framework implemented in {\tt{Zygote.jl}} \citep[see][]{Innes2019_Zygote_dP} and the {\tt{DiffEqFlux.jl}}-package \citep[see][]{Rackauckas2019}). 

In the scenario with four unknown ODE parameters, we employed a discretization to efficiently approximate the $L^2$-norm to determine the distance matrix with $m+1=11$ equidistant values by evaluating the ODE solutions at a fixed grid of $m+1$ time points $t_0,\dots, t_m$:
\begin{equation*}
d_{i,j} = \left( \frac{1}{m} \sum_{t=t_0}^{t_m} \Vert \mu_i(t) - \mu_j(t)\Vert_2^2 \right)^{\frac{1}{2}}.
\end{equation*}
Here, as an additional centralizing step, the mean of all $m+1$ values from the discretization is subtracted from each value to ensure that the distance is based on the actual development over the time interval, and not on random similarity of absolute values due to the necessary discretization.

\subsection*{A.1.\enspace Computational cost}

In the following, we provide details on the computational cost and runtimes of the numerical experiments. All results were obtained with an Intel Xeon-X5570 CPU running at $2.93$GHz with $70$ GiB of RAM.

Specifically, we repeated the fitting procedure of the model from Section \ref{subsec:sim-linear2parameters} with a linear underlying ODE system with two unknown parameters for varying numbers of observations, time-dependent variables and baseline variables. In Table \ref{tab:individuals}, we show the runtimes and memory consumption for the approach with $10$ time-dependent variables and $50$ baseline variables (the numbers used in all simulation scenarios) for varying numbers of individuals with observations at two time points each. 
Table \ref{tab:timedepvars} shows the runtimes and memory for the approach with $100$ individuals with observations at two time points each and $50$ baseline variables for varying numbers of time-dependent variables.
Finally, we report the runtimes and memory usage of the approach with $100$ individuals with observations at two time points each and $10$ time-dependent variables for varying numbers of baseline variables. Taken together, times and required memory scale roughly linearly with respect to the number of individuals and approximately exponentially with respect to the number of variables, with a slower increase for the baseline variables than for the time-dependent variables. 

\begin{table}[]
	\begin{tabular}{c|c|c}
		\textbf{Number of individuals} & \textbf{Time in seconds} & \textbf{Memory in GiB} \\ \hline
		50                             & 7.513                    & 1.58                   \\
		100                            & 14.972                   & 3.15                   \\
		250                            & 38.465                   & 7.88                   \\
		500                            & 71.82                    & 15.76                  \\
		1000                           & 145.507                  & 31.52                  \\
		2000                           & 278.739                  & 63.04                  \\
		5000                           & 704.397                  & 157.62                
	\end{tabular}
	\caption{Runtimes and memory usage of model fitting for a model based on a linear ODE system with two unknown parameters for varying numbers of individuals with observations at two time points each and fixed numbers of $10$ time-dependent variables and $50$ baseline variables.}
	\label{tab:individuals}
\end{table}

\begin{table}[]
	\begin{tabular}{c|c|c}
		\textbf{Number of time-dependent variables} & \textbf{Time in seconds} & \textbf{Memory in GiB} \\ \hline
		10                                          & 14.972                   & 3.15                   \\
		20                                          & 18.008                   & 3.65                   \\
		50                                          & 27.473                   & 5.53                   \\
		100                                         & 45.576                   & 9.91                   \\
		200                                         & 88.75                    & 23.33                 
	\end{tabular}
	\caption{Runtimes and memory usage of model fitting for a model based on a linear ODE system with two unknown parameters for varying numbers of time-dependent variables and fixed numbers of $100$ observations at two time points each and $10$ time-dependent variables.}
	\label{tab:timedepvars}
\end{table}

\begin{table}[]
	\begin{tabular}{c|c|c}
		\textbf{Number of baseline variables} & \textbf{Time in seconds} & \textbf{Memory in GiB} \\ \hline
		10                                    & 12.237                   & 3.08                   \\
		20                                    & 14.47                    & 3.09                   \\
		50                                    & 14.972                   & 3.15                   \\
		100                                   & 16.448                   & 3.36                   \\
		200                                   & 21.128                   & 4.17                  
	\end{tabular}
	\caption{Runtimes and memory usage of model fitting for a model based on a linear ODE system with two unknown parameters for varying numbers of time-dependent variables and fixed numbers of $100$ observations at two time points each and $50$ baseline variables.}
	\label{tab:baselinevars}
\end{table}

As described in Section \ref{subsec:background_integratingDEs}, to obtain gradients of the loss function with respect to the VAE encoder and the ODE-net, we need to differentiate backwards through the ODE solving step using automatic differentiation. Note that in our implementation, we achieve this by naively backpropagating gradients through the individual operations of the ODE solver. Alternatively, the adjoint sensitivity method could be used, which provides a more efficient solution with constant memory cost \citep{Chen2018}. While the naive method was sufficient for our computationally rather inexpensive experiments, the adjoint sensitivity method provides a more efficient alternative especially for larger models and more complex ODE systems, when computational cost is an issue. 

Regarding runtimes for the presented scenarios, it took $14.972$ seconds to train the model from Section \ref{subsec:sim-linear2parameters} based on a linear ODE system with two unknown parameters, $34.035$ seconds for the model from Section \ref{subsec:sim-nonlinear} based on a Lotka-Volterra ODE system with two unknown parameters, and $39.896$ seconds for the model from Section \ref{subsec:sim-linear4parameters} based on a linear ODE system with four unknown parameters using batches of similar individuals. 
Training on the SMArtCARE data as described in Section \ref{sec:SMArtCAREapplication} took $133.448$ seconds. 

\bibliographystyle{agsm}
\bibliography{bibfile.bib}

\end{document}